\documentclass[letterpaper,twocolumn,10pt]{article}
\usepackage{zhanggroup}

\usepackage{hyperref}
\hypersetup{
colorlinks,
linkcolor={blue!70!green},
citecolor={green!70!blue},
urlcolor={orange!70!red}
}
\usepackage{xspace}
\usepackage{times}
\usepackage{epsfig}
\usepackage{graphicx}
\usepackage{amsmath}
\usepackage{amssymb}
\usepackage{caption}
\usepackage{subcaption}
\usepackage{float}
\newcommand{\mypara}[1]{\smallskip\noindent{\bf {#1}.}}
\newcommand{\method}[1][]{\textsc{SeMap{#1}}\xspace}

\begin{document}

\date{}

\title{\Large \bf From Visual Prompt Learning to Zero-Shot Transfer: \\ Mapping Is All You Need}

\author{
Ziqing Yang \footnotemark[1] \ \ \ \ \
Zeyang Sha \footnotemark[1] \ \ \ \ \
Michael Backes\ \ \ \ \
Yang Zhang
\\
\\
\textit{CISPA Helmholtz Center for Information Security}
}

\maketitle

\renewcommand{\thefootnote}{\fnsymbol{footnote}}
\footnotetext[1]{The first two authors made equal contributions.}

\begin{abstract}

Visual prompt learning, as a newly emerged technique, leverages the knowledge learned by a large-scale pre-trained model and adapts it to downstream tasks through the usage of prompts.
While previous research has focused on designing effective prompts, in this work, we argue that compared to prompt design, a good mapping strategy matters more.
In this sense, we propose \method, a more effective mapping using the semantic alignment between the pre-trained model's knowledge and the downstream task.
Our experimental results show that \method can largely boost the performance of visual prompt learning.
Moreover, our experiments show that \method is capable of achieving competitive zero-shot transfer, indicating that it can perform the downstream task without any fine-tuning on the corresponding dataset.
This demonstrates the potential of our proposed method to be used in a broader range of applications where the zero-shot transfer is desired.
Results suggest that our proposed \method could lead to significant advancements in both visual prompt learning and zero-shot transfer.
We hope with \method, we can help the community move forward to more efficient and lightweight utilization of large vision models.

\end{abstract}

\section{Introduction}

\begin{figure}[t]
\begin{center}
\includegraphics[width=0.85\linewidth]{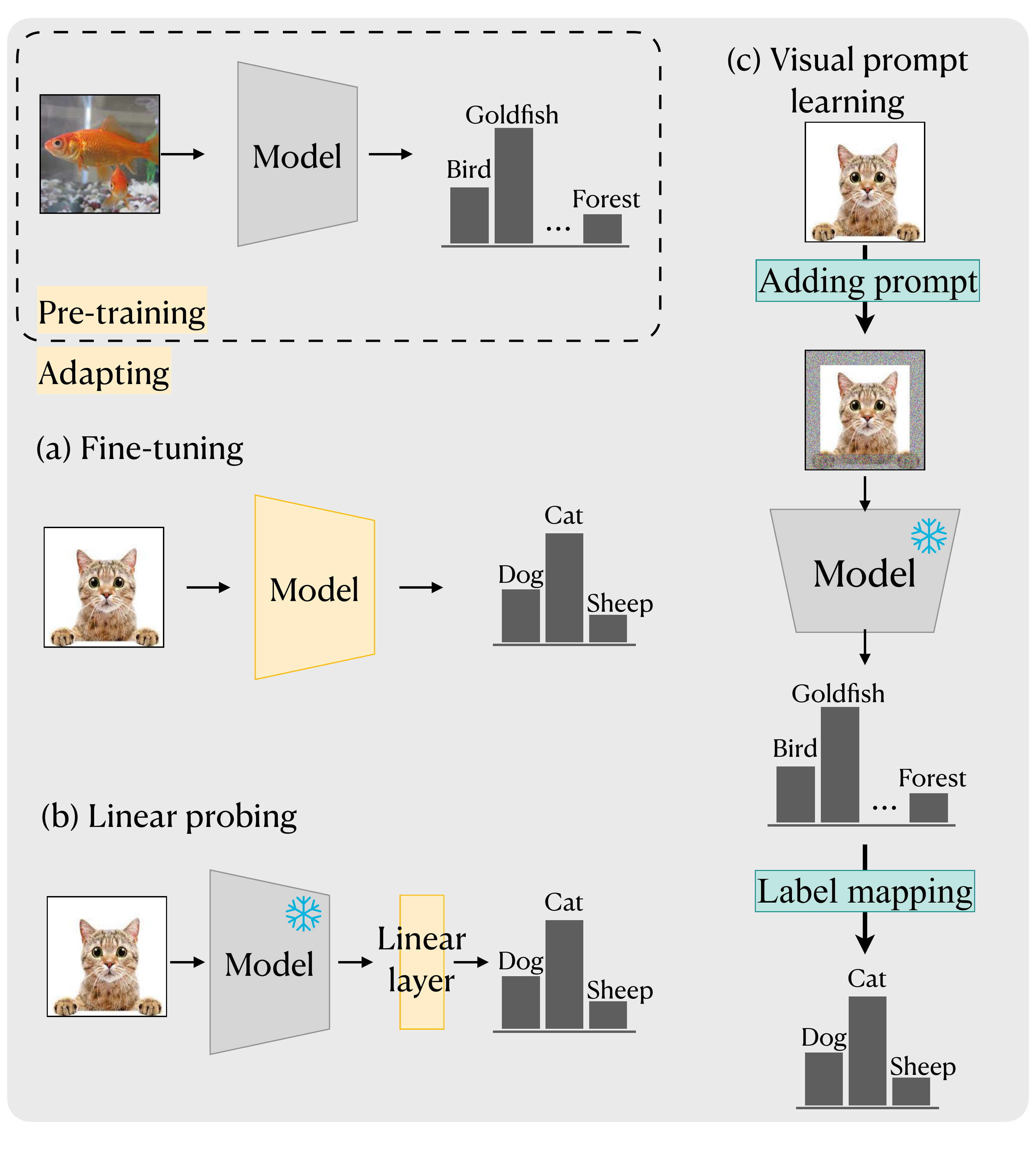}
\end{center}
\caption{
Overview of methods adapting the pre-trained model to downstream tasks.
(a) illustrates fine-tuning, which optimizes the whole model based on the downstream dataset.
(b) demonstrates linear probing, which replaces the final layer of the pre-trained model with a new linear layer, then trains only the new layer on the downstream dataset.
(c) is visual prompt learning, which includes a prompt and a mapping strategy.
The prompt is optimized based on the downstream dataset while the backbone model is kept frozen.
}
\label{figure:overview}
\end{figure}

Recently, large-scale pre-trained models such as ViT~\cite{DBKWZUDMHGUH21} and CLIP~\cite{RKHRGASAMCKS21} have shown their superiority in many applications~\cite{ZSGCBGGLD19, HZLFF20, KTN20, BDW21, MHB21}.
As such large-scale models are typically trained on massive datasets, they capture profound knowledge which can be adapted to many downstream tasks.
As shown in \autoref{figure:overview}, fine-tuning~\cite{DISFHS20, LCYLRBS20} and linear probing~\cite{HCXLDG21, RKHRGASAMCKS21} are two commonly used methods for this adaptation.
Fine-tuning updates the parameters of the pre-trained model by training on the downstream datasets.
However, optimizing a large pre-trained model is computationally intensive.
Additionally, the fine-tuned model usually becomes too specific to the training data and does not generalize well to new data.
Linear probing involves replacing the final layer of the pre-trained model with a new, untrained linear layer, and then training only the parameters of the new layer on the downstream dataset.
As only the new linear layer is optimized, linear probing adapts the pre-trained model to new tasks more efficiently.
However, both methods require some level of modifications to the pre-trained model, e.g., modifying some parameters or layers.

To overcome such limitations, visual prompt learning~\cite{EGS19, TMCEVH21, JTCCBHL22, BJSI22} has emerged as an efficient and effective alternative to fine-tuning for pre-trained vision models.
As shown in \autoref{figure:overview}~(d), visual prompt learning keeps the pre-trained model frozen, then learns a perturbation (i.e., a \emph{prompt}) added to the input image and maps the output of the pre-trained model to fit the downstream task.
Such a paradigm enables visual prompt learning to achieve efficient and lightweight knowledge transfer from pre-trained models.
However, existing methods~\cite{BJSI22, JTCCBHL22} only focus on how to design and optimize the prompt, neglecting the importance of the mapping strategy, i.e., mapping the output of the pre-trained model to the downstream task.
More importantly, in practice, the performance of visual prompt learning varies with different downstream tasks and is relatively low on many datasets.

In this paper, instead of focusing on designing proper prompts, we find that the mapping strategies matter more regarding boosting the performance of visual prompt learning.
More specifically, we propose \method (\textbf{Se}mantics-based \textbf{Map}ping), which explores the semantic alignment between the pre-trained task and the downstream tasks.
We further design \method[-1] and \method[-a], respectively, to enhance the capability of visual prompt learning concretely.
In \method[-1], we follow the intuition that there exist implicit connections between the output of the pre-trained model and the downstream task.
We try to map the semantically closest categories from the pre-trained task to downstream classes.
To achieve this, we leverage the text encoder of CLIP~\cite{RKHRGASAMCKS21} to generate text embeddings of both pre-trained class labels and downstream class labels.
Then, we measure the semantic similarity by calculating the cosine similarity between two embeddings.
Thus the downstream class would be mapped to its most similar class in the pre-trained task.

Unlike 1-on-1 mapping in \method[-1], \method[-a] tries to map multiple indices to one downstream class to leverage the pre-trained model's information better, i.e., an $k$-on-1 mapping.
To find the best $k$ for different downstream classes, we propose an adaptive method \method[-a].
Our adaptive method can find the most representative $k$ based on the ideology of clustering.
Given a downstream class, our algorithm will first sort the similarities of all pre-trained classes.
We further find a gap between different similarities and then divide the pre-trained classes into top-$k$ classes and other classes.
Indices correspond to top-$k$ classes will be mapped to the given downstream class.

To evaluate the effectiveness and efficiency of our proposed \method, we first compare it with existing visual prompt learning~\cite{CFY21, BJSI22}.
Our evaluation results demonstrate that with \method, visual prompt learning can have much better performance than previous work.
For instance, methods proposed by Bahng et al.~\cite{BJSI22} can only achieve 0.412 accuracy on CIFAR10~\cite{CIFAR} ResNet18~\cite{HZRS16}, while \method (including \method[-1] and \method[-a]) can achieve 0.570/0.642, which is much higher.
We also find that even the downstream datasets are totally different from pre-trained datasets, e.g., Fashion-MNIST~\cite{XRV17}, \method based on semantic meaning mapping can still have better performance than just random map in previous works.
Apart from visual prompt learning, we further point out that with the mapping strategies in \method, we can also achieve great zero-shot transfer performance without training the prompts.
For instance, with \method[-a], the zero-shot transfer can have 0.658 accuracy on STL10~\cite{STL10} ResNet50~\cite{HZRS16}, which is even higher than the visual prompt learning methods proposed in Bahng et al~\cite{BJSI22}.

Our contributions can be summarized as follows.
\begin{itemize}
\item We are the first to systematically summarize the existing visual prompt learning on pre-trained image classifiers.
\item We are the first to demonstrate that in visual prompt learning, mapping strategies matter more than the design of prompts.
We further proposed \method, a mapping method based on semantics.
\item Extensive experiments show that \method can largely boost the performance of visual prompt learning.
\item \method itself can achieve great zero-shot transfer performance, and its effectiveness and efficiency are shown in many downstream tasks.
\end{itemize}

\section{Related Work}

\subsection{Prompt Learning}

\mypara{Prompt Learning in NLP}
Prompt learning~\cite{LYFJHN21} has been extensively studied and applied in the field of natural language processing (NLP).
Take a pre-trained language model as the backbone, a prompt is a task-specific template used to reformulate the downstream task's input.
Consider the task of predicting the emotion (positive/negative) of a given text.
A suitable prompt for this task might be ``[INPUT] All in all, I felt [MASK].''
Given the input ``The sunshine is warm and beautiful,'' the backbone model might fill the mask as ``wonderful'', which conveys a positive emotion, enabling the output to be classified accordingly.
Hand-crafting an effective textual prompt for a downstream task can improve prompt learning.
However, this process often requires significant domain expertise and a great deal of trial-and-error experimentation.
AutoPrompt~\cite{SRIWS20} is the first to study how to automatically get the appropriate prompt from the vocabulary.
However, determining proper words is discrete and hard to optimize.
Recent approaches~\cite{HKM21, LL21, LAC21, LZDDQYT21, HZDLS22, BOR22} aim to address these issues by focusing on learning a "soft prompt," which is a continuous vector that can be optimized via backpropagation while keeping the backbone model frozen.
Notable examples of this approach include prompt tuning~\cite{LL21} and hybrid tuning~\cite{LZDDQYT21}.

\mypara{Prompt Learning in CV}
Building on the success of prompt learning in NLP, researchers have explored the possibility of adapting this approach to the domain of computer vision (CV), resulting in the emergence of visual prompt learning.
Before prompt learning, researchers identified an attack known as adversarial reprogramming in computer vision.
Adversarial reprogramming~\cite{EGS19} is an adversarial attack that repurposes pre-trained models without modifying any of their parameters by adding a perturbation to the input image.
Despite having different motivations, the input perturbation acts as a prompt.
Enlightened by this, works on visual prompt learning~\cite{CFY21, JTCCBHL22, BJSI22} are proposed, aiming at transferring the knowledge of the pre-trained model to downstream tasks.
For example, Chen et al.~\cite{CFY21} further enhance adversarial reprogramming as a transfer learning strategy.
And they propose a frequency-based mapping strategy.
Jia et al.~\cite{JTCCBHL22} proposes VPT that is specific to Vision Transformers~\cite{DBKWZUDMHGUH21}.
Bahng et al.~\cite{BJSI22}, instead, focus on pre-trained vision models including image classification models and CLIP~\cite{RKHRGASAMCKS21} models, and propose a prompt learning framework.
The prompt is an image perturbation learned via backpropagation while keeping the backbone model frozen.
However, they only use the random mapping strategy, which maps the top indices of the output to the downstream task's labels.
In this paper, we will follow this framework and mainly investigate visual prompt learning on image classification models.

\subsection{Zero-Shot Transfer}

The concept of zero-shot transfer is first proposed by Radford et al.~\cite{RKHRGASAMCKS21}.
Zero-shot transfer is a method that aims to perform unseen tasks by utilizing the knowledge learned from related tasks, \emph{without} additional training or fine-tuning.
It focuses on measuring the task learning capabilities of machine learning models.
Different from zero-shot learning, zero-shot transfer emphasizes without training or fine-tuning on ``unseen'' tasks.
Recently, there are also some extensions of zero-shot transfer~\cite{JYXCPPLSLD21, ZWMSKKB22, CWCPPSGGMBKPDRAMXTBKSJARSAZHS22}.

\section{Preliminaries}

\begin{figure*}[t]
\begin{center}
\includegraphics[width=0.8\linewidth]{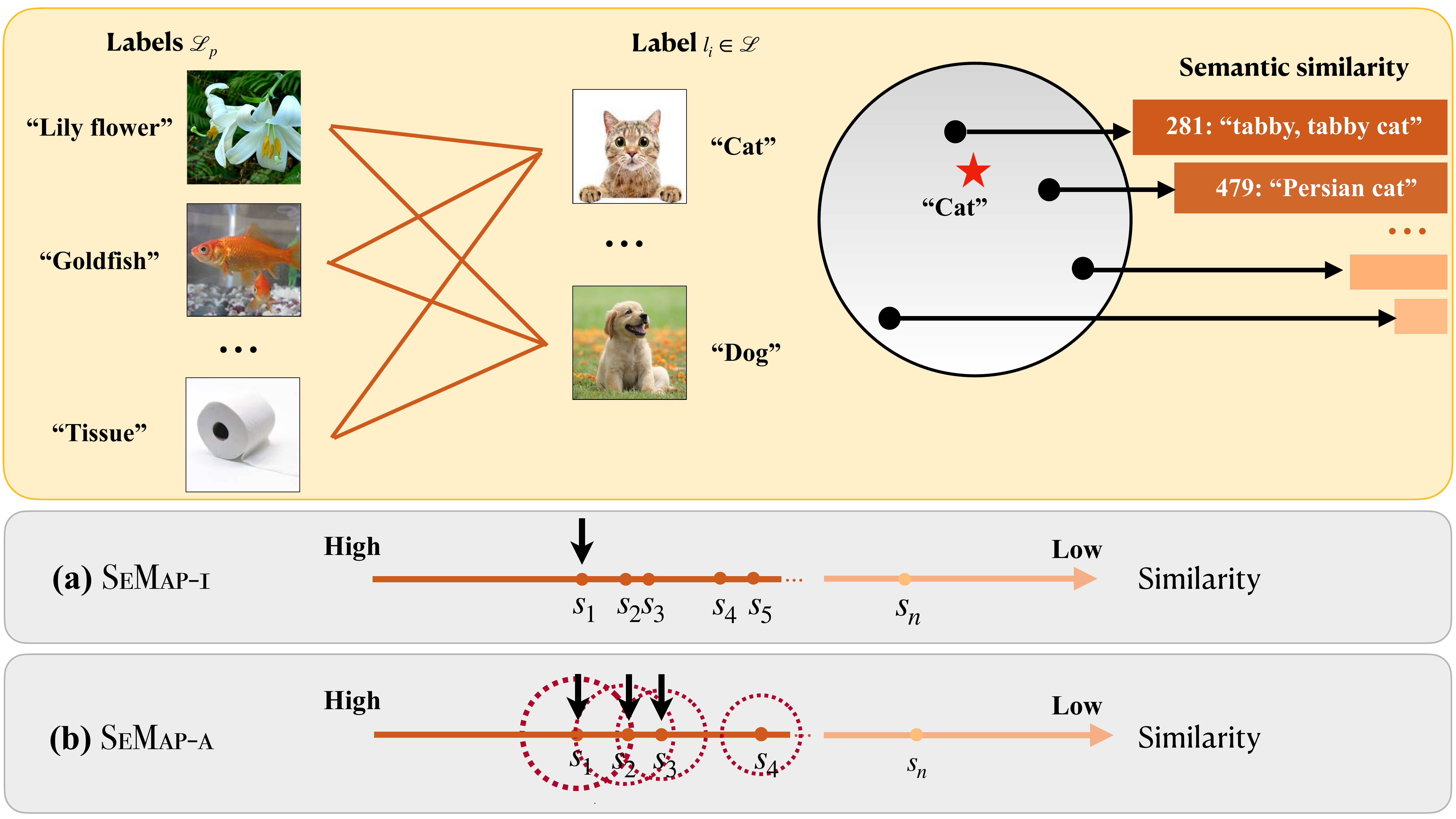}
\end{center}
\caption{
Overview of \method, which is based on semantic similarity.
For each label $l^i \in \mathcal{L}$, we compute its semantic similarity with all labels $\mathcal{L}_p$ in the pre-trained task.
We then sort the indices by their similarities from high to low.
Specifically, (a) \method[-1] maps the index with the highest similarity to the $i$-th class.
(b) \method[-a] is an adaptive $k$-on-1 mapping method that defines $k$ for each label automatically by introducing a hyperparameter $\epsilon$.
}
\label{figure:semap}
\end{figure*}

An image classification model is a parameterized function $f_{\Theta}: \mathbb{R}^{d\times d} \to \mathbb{R}^n$ that maps a $d\times d$ image $x$ to an output $y \in \mathbb{R}^n$.
Here, $\Theta$ represents the parameters of the function $f$, $d$ is the size of the image, and $n$ corresponds to the number of classes to be predicted.
More specifically, the output $y$ is a probability distribution over $n$ possible classes.
The label of the image is predicted as the class with the highest probability, i.e., $\arg \max_{i \in [1, n]}{y^i}$.

\section{Mapping Is All You Need}

In this section, we first systematically summarize the framework of visual prompt learning.
Then we propose our method \method as the mapping strategy to enhance visual prompt learning.
We also show that our proposed \method can adapt large pre-trained visual models to new downstream tasks without \emph{any} optimization, i.e., can achieve zero-shot transfer in downstream tasks.

\subsection{Visual Prompt Learning}

We first define how visual prompt learning works.
The framework consists of a backbone model $f_{\Theta}$, a visual prompt $\Delta$, and a mapping strategy $h$.

\mypara{Backbone Model}
The backbone model is an image classification model $f_{\Theta}$ pre-trained on a pre-trained dataset $(\mathcal{X}_p, \mathcal{Y}_p)$, where $\mathcal{X}_p \subset \mathbb{R}^{d\times d}$ contains images, and $\mathcal{Y}_p \subset \mathbb{R}^n$ corresponds to the classes.

\mypara{Visual Prompt}
The visual prompt is normally a trainable perturbation $\Delta$ added to an image $x$ so that images with that perturbation in the downstream dataset $(\mathcal{X}, \mathcal{Y})$ can be adapted to the input space of the backbone model, where $\mathcal{X} \subset \mathbb{R}^{d\times d}$ and $\mathcal{Y} \subset \mathbb{R}^m$.
Specifically, we can present the prompted image as $\oplus(x, \Delta) \in \mathbb{R}^{d\times d}$, where $\oplus(\cdot, \cdot)$ represents adding the perturbation to the image.
The prompted image has the capability of spurring the backbone model to perform a certain downstream task.
As discussed in Bahng et al.~\cite{BJSI22}, $\Delta$ can be a pixel patch at a random/fixed location or padding around the image.
Chen et al.~\cite{CFY21} perturb the whole image so that $\Delta$ can also have the same size as the input image.
In this work, we focus on the prompt that is padding around the image as padding achieves the best performance over other design choices~\cite{BJSI22}.

\mypara{Mapping Strategy}
The mapping strategy is an output transformation that maps the output of the backbone model to the labels of the downstream task, i.e., $h(y_p)=y, y_p \in \mathbb{R}^n, y \in \mathbb{R}^m$.
There are several existing mapping strategies.
\begin{itemize}
\item \mypara{Random Mapping (RM)}
We start from this simplest mapping strategy that is used in many existing works~\cite{EGS19, CFY21, BJSI22}.
RM randomly selects $m$ indices from the output $y_p \in \mathcal{Y}_y$ and maps them to the output $y \in \mathcal{Y}$ of the downstream task
One simplest way is to map the top $m$ indices of $y_p$ to $y$.
It can be written as:
\begin{equation}
y^i = y_p^i, i \in [1,m].
\end{equation}
The remaining $n-m$ indices of $y_p$ are abandoned.
\item \mypara{Frequency-Based Mapping (FM)}
Although RM is simple and easy to conduct, such a strategy ignores the connection between the pre-trained task and the downstream task.
Chen et al.~\cite{CFY21} propose FM, which makes use of the information contained in the backbone model.
Suppose $\mathcal{X}^i$ denotes the images in class $i$ in $\mathcal{X}$.
For $x \in \mathcal{X}^i$, we can get $y_p=f_{\Theta}(x)$, which is the original output without prompt.
And the class index of the pre-trained task it corresponds to is $\arg \max_{j \in [1,n]}{y_p^j}$.
Based on the outputs of all $x \in \mathcal{X}^i$, we can assign the most frequent class index $j$ to the corresponding class $i$ in the downstream task.
For each $y^i \in y, i \in [1,m]$, it is mapped to $y_p^j \in y_p, j\in [1,n]$ based on frequency.
And $j$ is defined as the most frequent class index for images in class $i$ in the downstream dataset.
\begin{equation}
j=\arg \max_{j\in [1,n]}{\mathbb{E}(f_{\Theta}{(x)} \text{ is } j| x \in \mathcal{X}^i)}.
\end{equation}
\end{itemize}

\mypara{Overall}
As demonstrated in \autoref{figure:overview}~(c), visual prompt learning first adds a prompt $\Delta$ on the input image and then puts it into the backbone model $f_{\Phi}$, finally, the output is mapped by the mapping strategy $h$.
Based on the training dataset $(\mathcal{X}, \mathcal{Y})$, visual prompt learning optimizes the prompt $\Delta$ via backpropagation.
It aims to optimize:
\begin{equation}
\arg\min_{\Delta} \mathbb{E}_{(x,y)\in (\mathcal{X}, \mathcal{Y})}{L(h\circ f \circ \oplus(x,\Delta), y)}.
\end{equation}
Here, $L(\cdot, \cdot)$ denotes the cross-entropy loss, which is widely used in supervised learning.

We observe that current mapping strategies have some limitations.
As shown above, RM randomly assigns the indices, ignoring any connection between the pre-trained and downstream tasks.
As for FM, it maps the most frequent index to the corresponding class.
However, it requires obtaining a set of examples first.
For each class, FM first requires some images in this class, then obtains their outputs from the model and computes the frequency.
Therefore, FM is time-consuming and requires access to the downstream dataset before conducting visual prompt learning.

\subsection{\method}

In this paper, we propose \method, which only requires the label information of the pre-trained and downstream tasks instead of access to the downstream dataset.
We now discuss how our proposed \method works.
We divide our method into two parts, i.e., \method[-1] and \method[-a].
\method[-1] adopts the intuitive way to conduct the map, e.g., finding the class of the pre-train dataset with the closest semantics of the downstream class.
\method[-a] is an adaptive method, which adopts $k$-on-1, instead of 1-on-1, to map multiple classes to one downstream class.
\method[-a] aims to find the best $k$ for each downstream class automatically.
We will introduce these two methods below.

\mypara{\method[-1]}
We start with a simple setting that we map one index of the backbone model's output to each downstream class.
Recall that the pre-trained dataset and the downstream dataset are labeled.
We denote the labels of the pre-trained dataset as $\mathcal{L}_p$ and that of the downstream dataset as $\mathcal{L}$.
Each label is a natural language description that contains the semantic information of the corresponding class.
To gain a better mapping strategy, we make use of the semantic alignment between $\mathcal{L}_p$ and $\mathcal{L}$.
For each label $l^i \in \mathcal{L}$, we measure its semantic similarity with all labels in $\mathcal{L}_p$.
Then we sort the indices by the similarity from high to low as $\mathcal{S} = \{s_1, s_2, \cdots, s_n\}$.
For example, for a label $l^i \in \mathcal{L}$, the most similar label would be $l_p^{s_1}$, as shown in \autoref{figure:semap}~(a).
In other words, $s_1=\arg \max_{j \in [1,n]}{\text{Sim}(l_p^j, l^i)}$, where $\text{Sim}(\cdot, \cdot)$ represents the semantic similarity.
Naturally, based on the semantic similarity, \method maps the index with the most similar label to the target downstream class:
\begin{equation}
y^i = y_p^{s_1}, i \in [1,m].
\end{equation}
We first put the label descriptions into the text encoder of the CLIP model~\cite{RKHRGASAMCKS21}, then the semantic similarity is given by computing the cosine similarity between two text embeddings.
The reason we choose CLIP is that CLIP learns the alignment between texts and images so that its text and image embeddings are in the same space.
We notice that there could be one label that is the most similar one with several downstream labels.
For example, ``bird'' and ``airplane'' in STL110 are both quite similar to ``fly'' in ImageNet-1k.
In this case, \method[-1] would find the next similar label and assign its index to the downstream label.

\mypara{\method[-a]}
Although \method[-1] successfully aligns the semantic meaning, it only takes advantage of limited information from the output of the backbone model.
Therefore, we develop \method[-a] to overcome such drawbacks.
\method[-a] generalizes \method[-1] to a multi-label mapping setting, i.e., we map multiple indices of the model's output to one class in $\mathcal{Y}$.
We define $k$ as the number of multiple indices mapped to one class.

\begin{figure*}[!t]
\centering
\begin{subfigure}{0.5\columnwidth}
\includegraphics[width=\columnwidth]{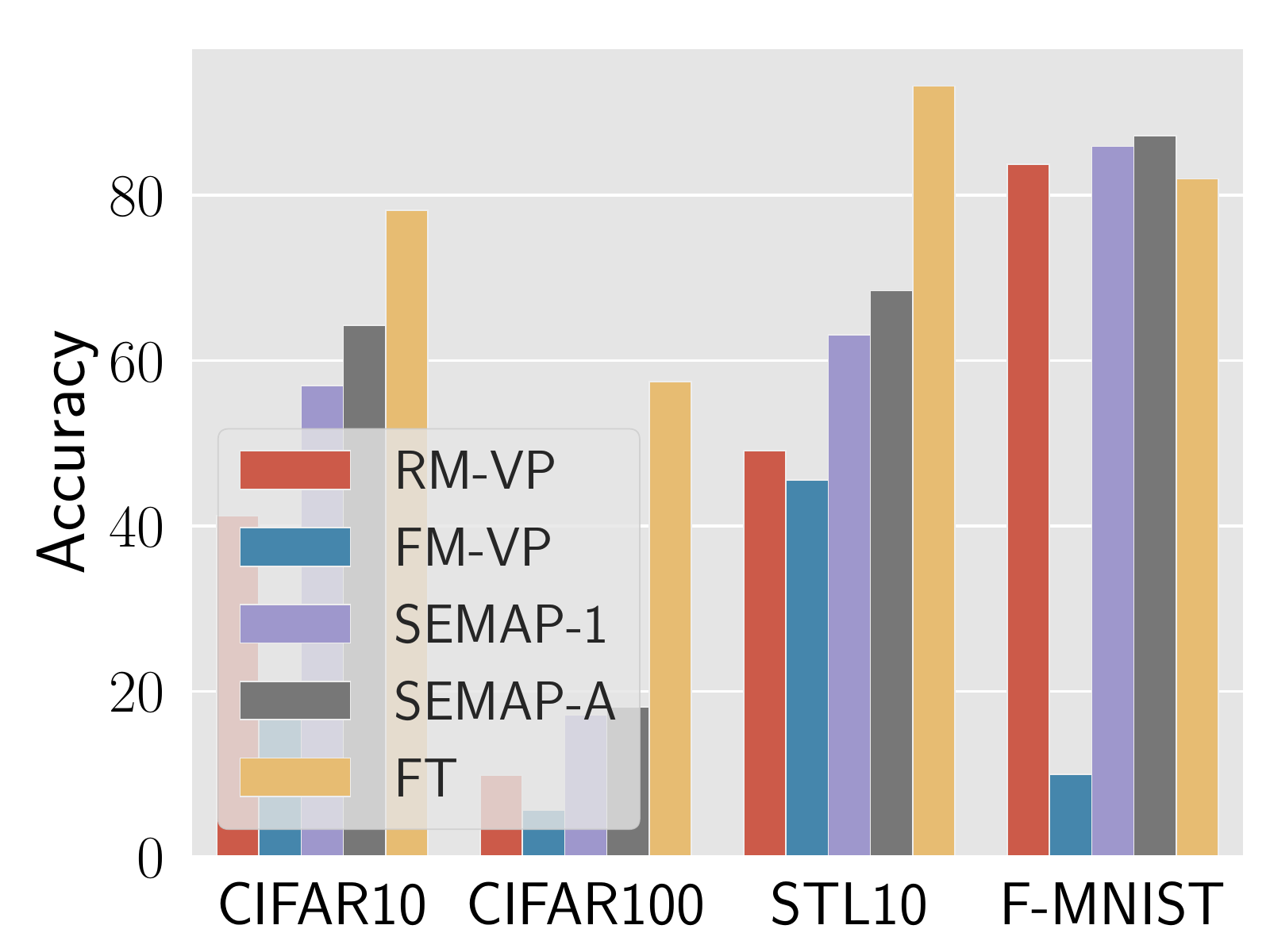}
\caption{ResNet18}
\label{figure:rn18_vp}
\end{subfigure}
\begin{subfigure}{0.5\columnwidth}
\includegraphics[width=\columnwidth]{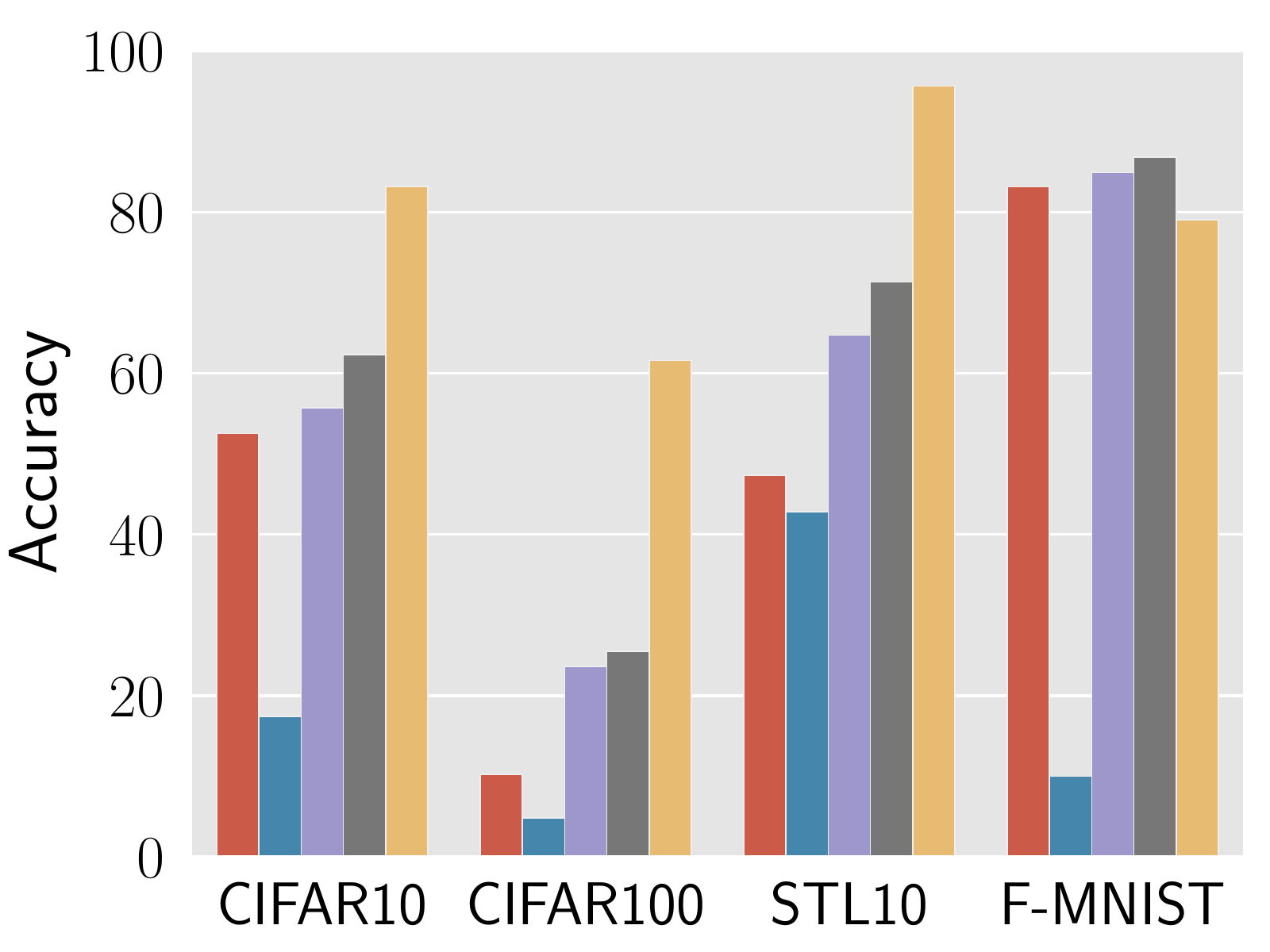}
\caption{ResNet50}
\label{figure:rn50_vp}
\end{subfigure}
\begin{subfigure}{0.5\columnwidth}
\includegraphics[width=\columnwidth]{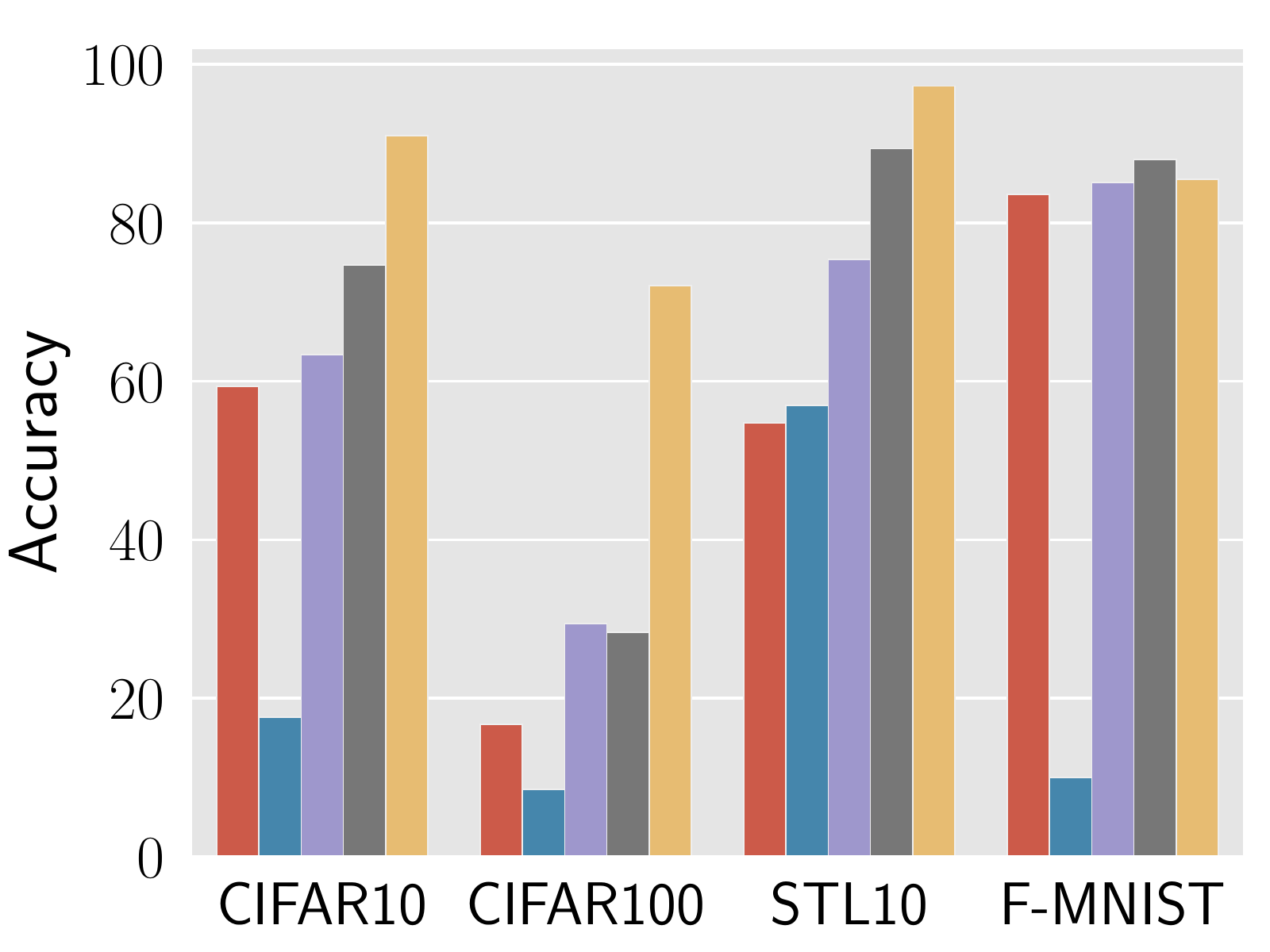}
\caption{BiT-M ResNet50}
\label{figure:bit_vp}
\end{subfigure}
\begin{subfigure}{0.5\columnwidth}
\includegraphics[width=\columnwidth]{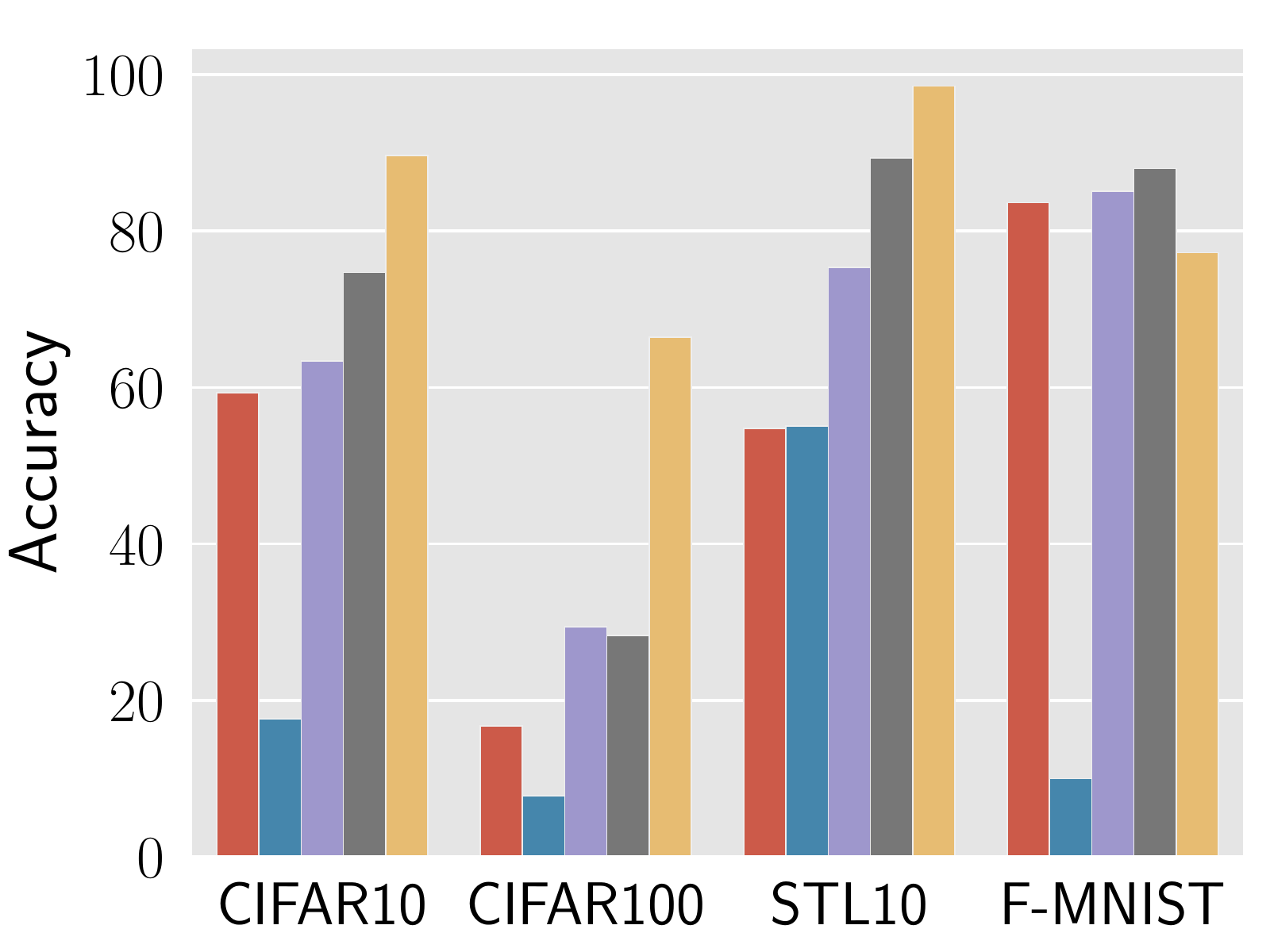}
\caption{Instagram ResNeXt}
\label{figure:ins_vp}
\end{subfigure}
\caption{The performance of RM-VP, FM-VP, \method[-1], \method[-a], and FT on ResNet18, ResNet50, Instagram ResNeXt ,and BiT-M ResNet50.
The x-axis represents different downstream datasets (CIFAR10, CIFAR100, STL10, and F-MNIST).
The y-axis represents the test accuracy.
}
\label{figure:vp}
\end{figure*}

\begin{figure*}[!t]
\centering
\begin{subfigure}{0.5\columnwidth}
\includegraphics[width=\columnwidth]{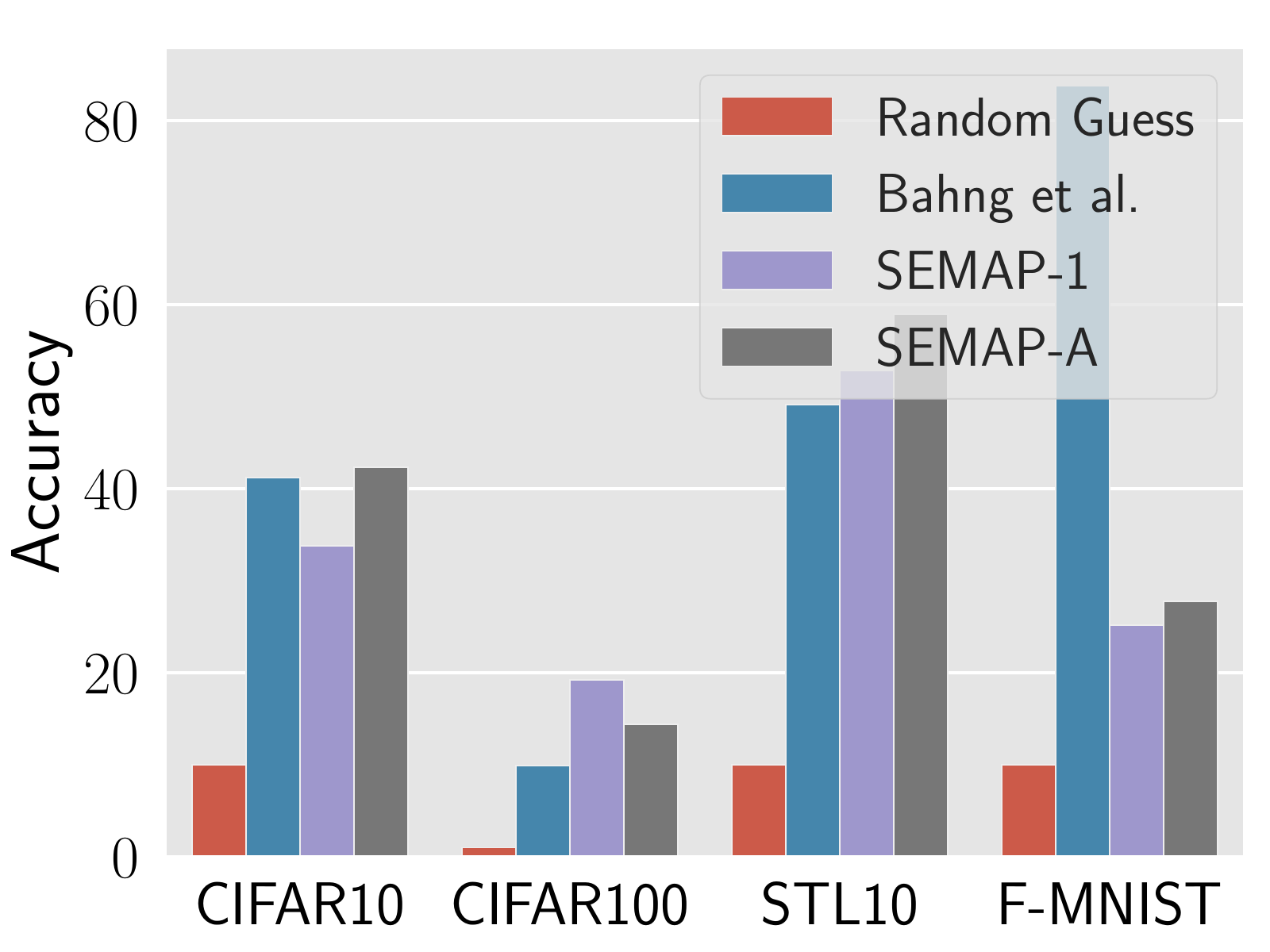}
\caption{ResNet18}
\label{figure:rn18_vp-zero-shot}
\end{subfigure}
\begin{subfigure}{0.5\columnwidth}
\includegraphics[width=\columnwidth]{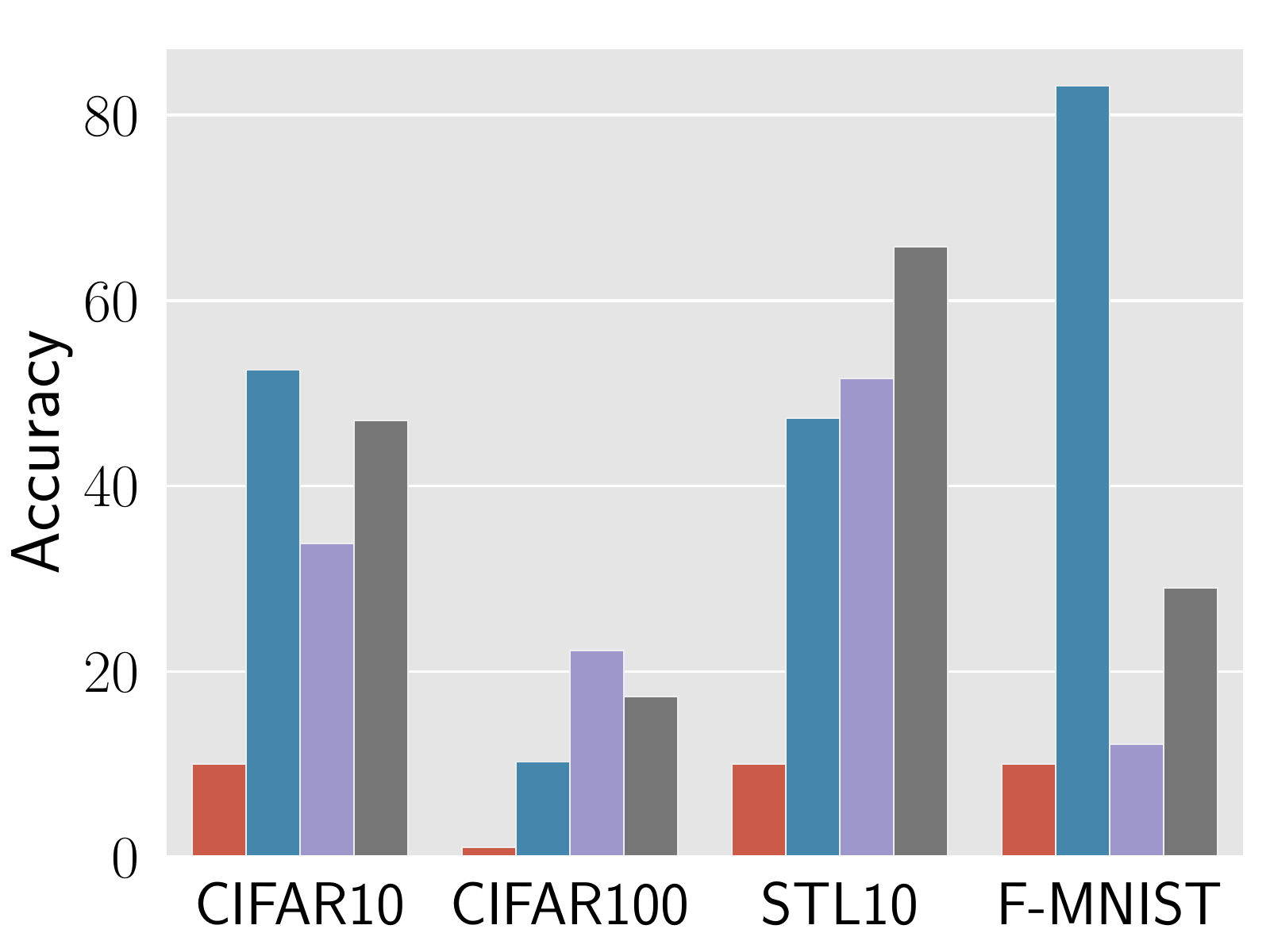}
\caption{ResNet50}
\label{figure:rn50_vp-zero-shot}
\end{subfigure}
\begin{subfigure}{0.5\columnwidth}
\includegraphics[width=\columnwidth]{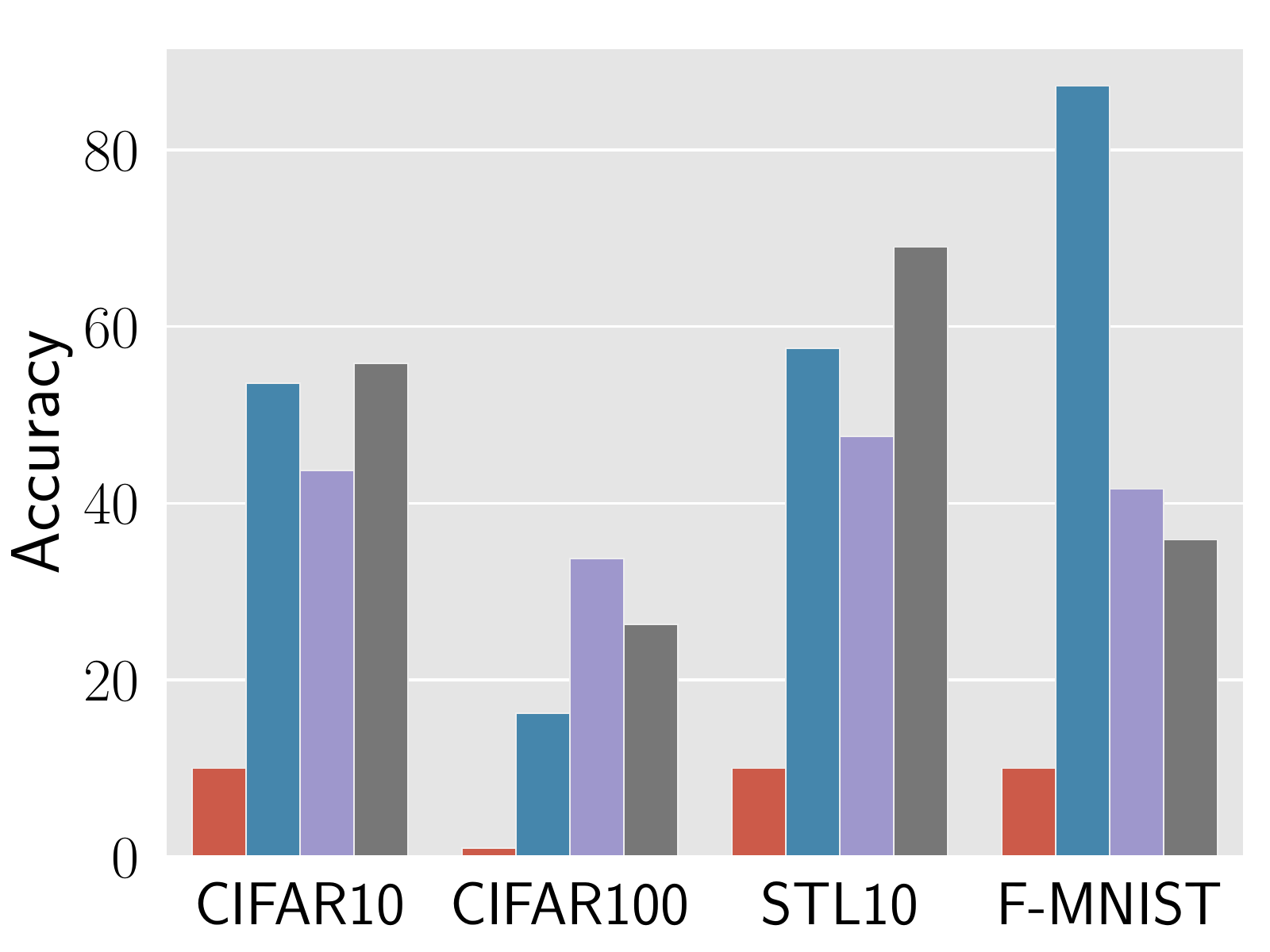}
\caption{BiT-M ResNet50}
\label{figure:bit_vp-zero-shot}
\end{subfigure}
\begin{subfigure}{0.5\columnwidth}
\includegraphics[width=\columnwidth]{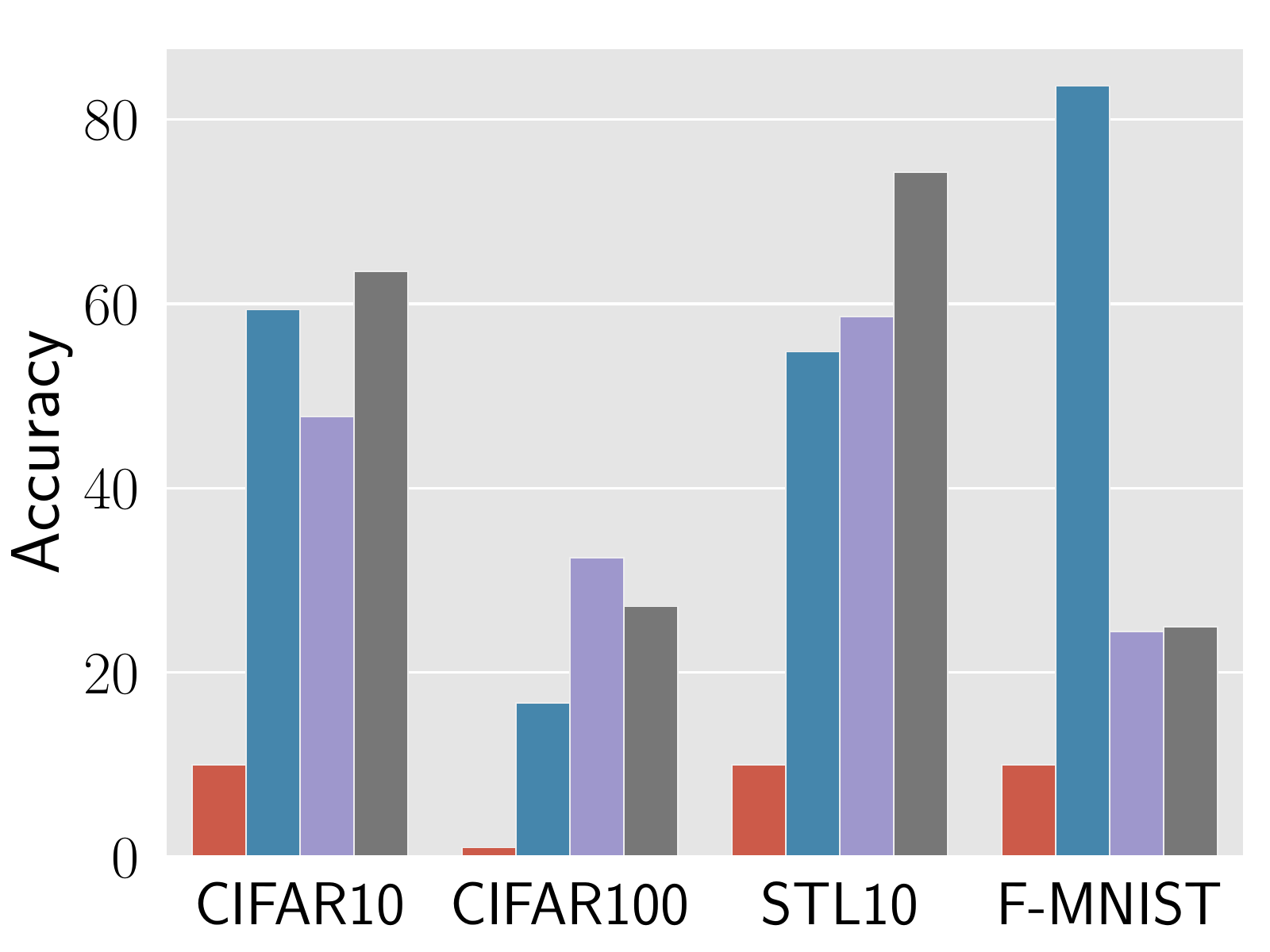}
\caption{Instagram ResNeXt}
\label{figure:ins_vp-zero-shot}
\end{subfigure}
\caption{The performance of zero-shot transfer of Random Guess, \method[-1], \method[-a], and RM-VP on ResNet18, ResNet50, Instagram ResNeXt ,and BiT-M ResNet50.
The x-axis represents different downstream datasets (CIFAR10, CIFAR100, STL10, and F-MNIST).
The y-axis represents the test accuracy.}
\label{figure:zero-shot}
\end{figure*}

Different from defining a unified $k$ for all classes, \method[-a] aims at automatically assigning a $k$ for each label $l^i \in \mathcal{L}$.
We first introduce two concepts named \emph{reachability} and \emph{connectivity}.
Recall that for a label $l^i \in \mathcal{L}$, the corresponding indices sorted by the similarity from high to low are $\mathcal{S} = \{s_1, s_2, \cdots, s_n\}$.
As shown in \autoref{figure:semap}, $\mathcal{S}$ can be considered as a set of points lying on a line, the value of each point is the corresponding semantic similarity.
For any two points in $\mathcal{S}$, they are \emph{reachable} if the distance between them, i.e., their semantic similarity difference, is smaller than a threshold.
Based on transitivity, two points are \emph{connected} if there exists a path that any two points next to each other are {reachable}.
For example, for $s_s, s_t \in \mathcal{S}$, if there exists a path/sequence $s_s \leftrightarrow s_{a} \leftrightarrow s_{b} \leftrightarrow s_{t}$ where $\leftrightarrow$ denotes reachable, then $s_s$ and $s_t$ are connected.

\method[-a] aims to find all points that are connected with $s_1$.
Here we introduce two hyperparameters, a distance threshold $\epsilon$ and a decay rate $\gamma$.
The distance $\epsilon$ denotes the initial distance threshold in \method[-a], i.e., two points whose distance is shorter than $\epsilon$ will be considered reachable.
After we go to the next point, the distance threshold will decrease by the decay rate $\gamma$.
For example, when we centered on $s_3$, the distance threshold would be ${\gamma}^2 \epsilon$.
In this way, the mapping function can be written as:
\begin{equation}
\label{equation:semap-a}
y^i = \sum_{j \in [1,k_i]}^{k}{y_p^{s_{j}}}.
\end{equation}
Here, $k_i$ is assigned for each label adaptively, which denotes the number of points that are connected with $s_1$.
Note that \method[-1] is the special case of \method[-a] when $k=1$.

\mypara{Computational Complexity}
\method is efficient from the perspective of computational complexity.
For a label $l^i \in \mathcal{L}$, the complexity for getting the index with the highest similarity is $O(\log n)$ using max heap, where $n$ corresponds to the number of pre-trained labels.
There are $m$ labels in $\mathcal{L}$, thus the time complexity of \method[-1] is $m\log n$, respectively.
As for \method[-a], for $l^i \in \mathcal{L}$, the worst case is to go through all points, i.e., $O(n\log n)$.
But we can bound the complexity by setting a maximum number $C$, i.e., after we get $C$ points, we will stop searching.
In this sense, the complexity for \method[-a] is $O(m*C\log n) = O(m \log n)$.

\subsection{From Visual Prompt Learning To Zero-Shot Transfer}

We further find that \method can also be adapted to zero-shot transfer settings.
As recently in-context learning in NLP~\cite{BMRSKDNSSAAHKHCRZWWHCSLGCCBMRSA20, MLHALHZ22} has been quite popular, the mapping process can be considered as providing the context of a given image for the pre-trained model.
Therefore, users can only use the mapping strategy we have proposed to adapt the pre-trained model to their own downstream tasks without any optimization.
The whole process can be described as 
\begin{equation}
y_{\text{predict}} = \arg \max_{j\in [1,m]}{(h \circ f_{\Theta} (x))^j}
\end{equation}
Here $h(\cdot)$ represents \method[-a] and is defined by \autoref{equation:semap-a}, and $f_{\Theta}(\cdot)$ denotes the pre-trained model, respectively.
We further demonstrate that with \method, zero-shot transfer can even have stronger performance than existing visual prompt learning works.

\section{Experiments}

\begin{figure*}[!t]
\centering
\begin{subfigure}{0.5\columnwidth}
\includegraphics[width=\columnwidth]{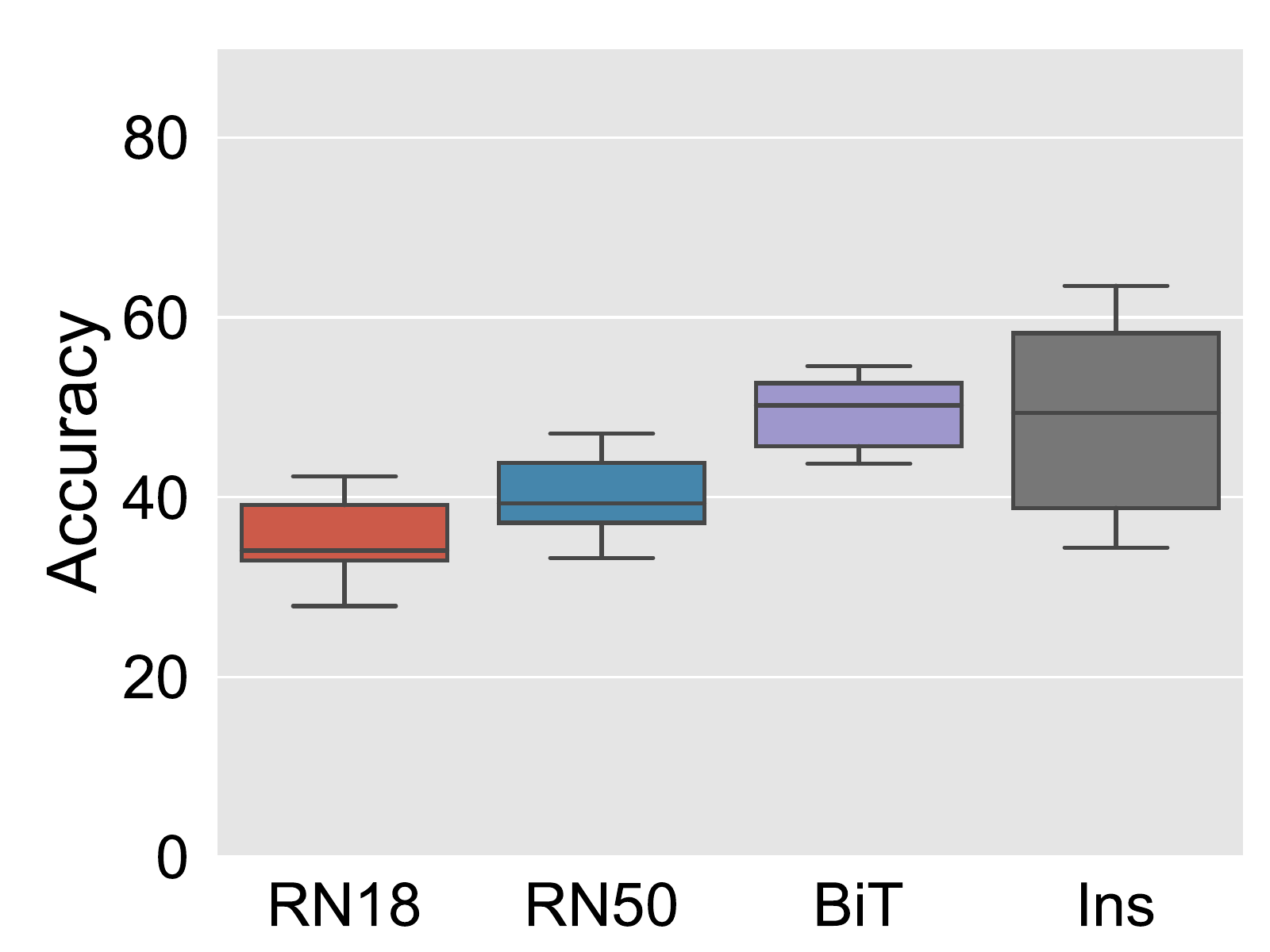}
\caption{Zero-Shot: CIFAR10}
\label{figure:ablation_k_cifar10}
\end{subfigure}
\begin{subfigure}{0.5\columnwidth}
\includegraphics[width=\columnwidth]{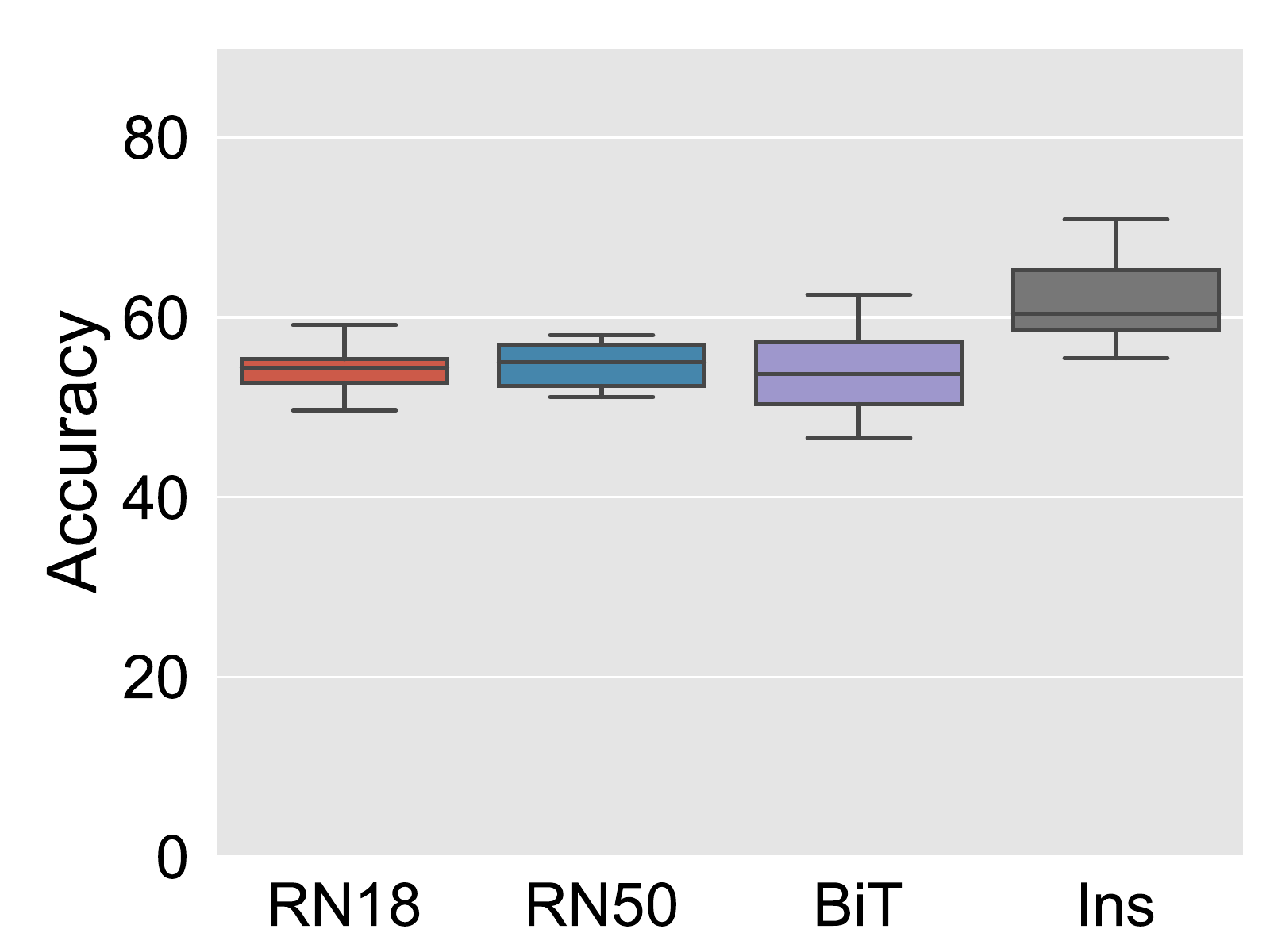}
\caption{VP: CIFAR10}
\label{figure:ablation_k_cifar10-prompt}
\end{subfigure}
\begin{subfigure}{0.5\columnwidth}
\includegraphics[width=\columnwidth]{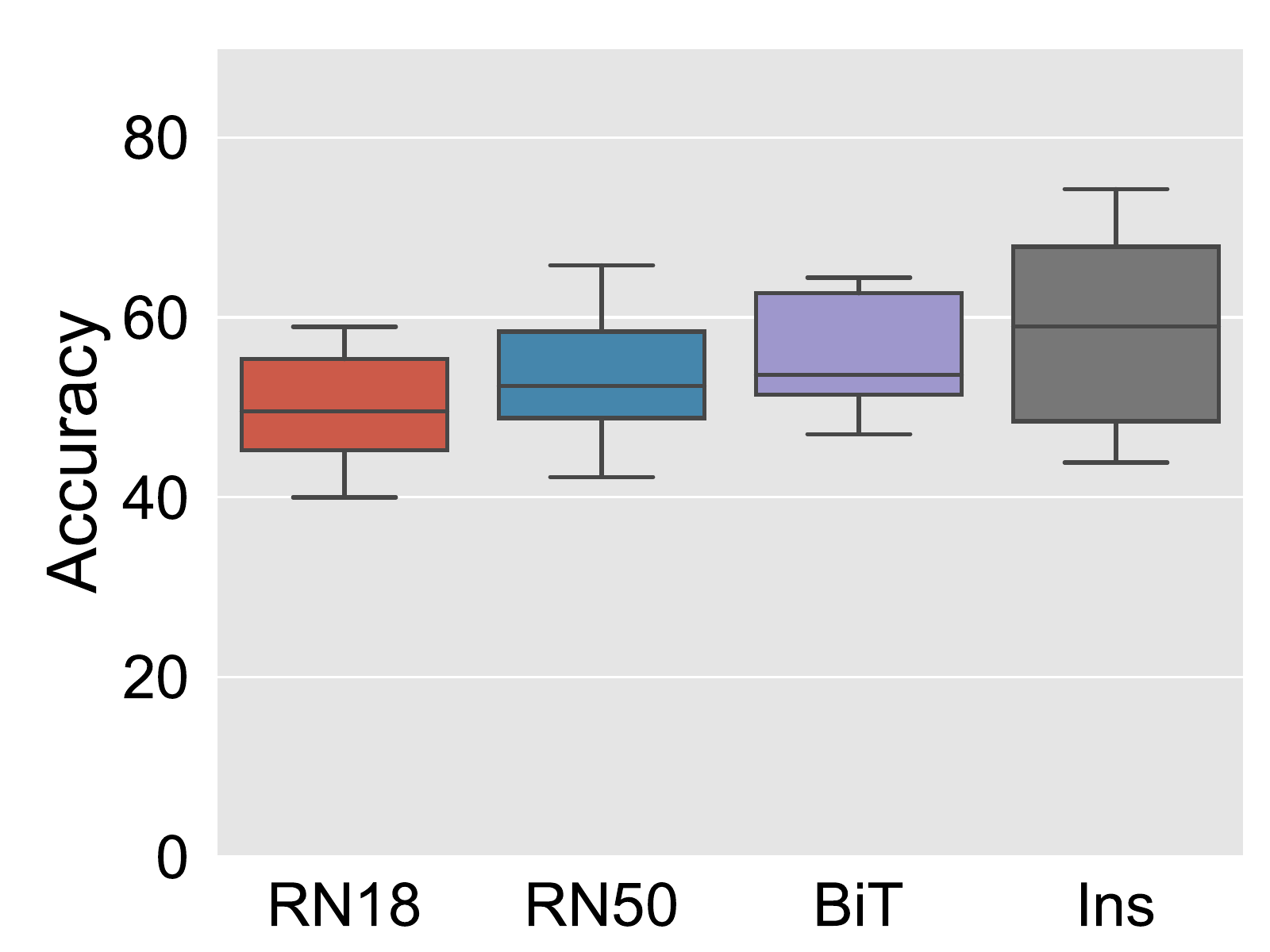}
\caption{Zero-Shot: STL10}
\label{figure:ablation_k_stl10}
\end{subfigure}
\begin{subfigure}{0.5\columnwidth}
\includegraphics[width=\columnwidth]{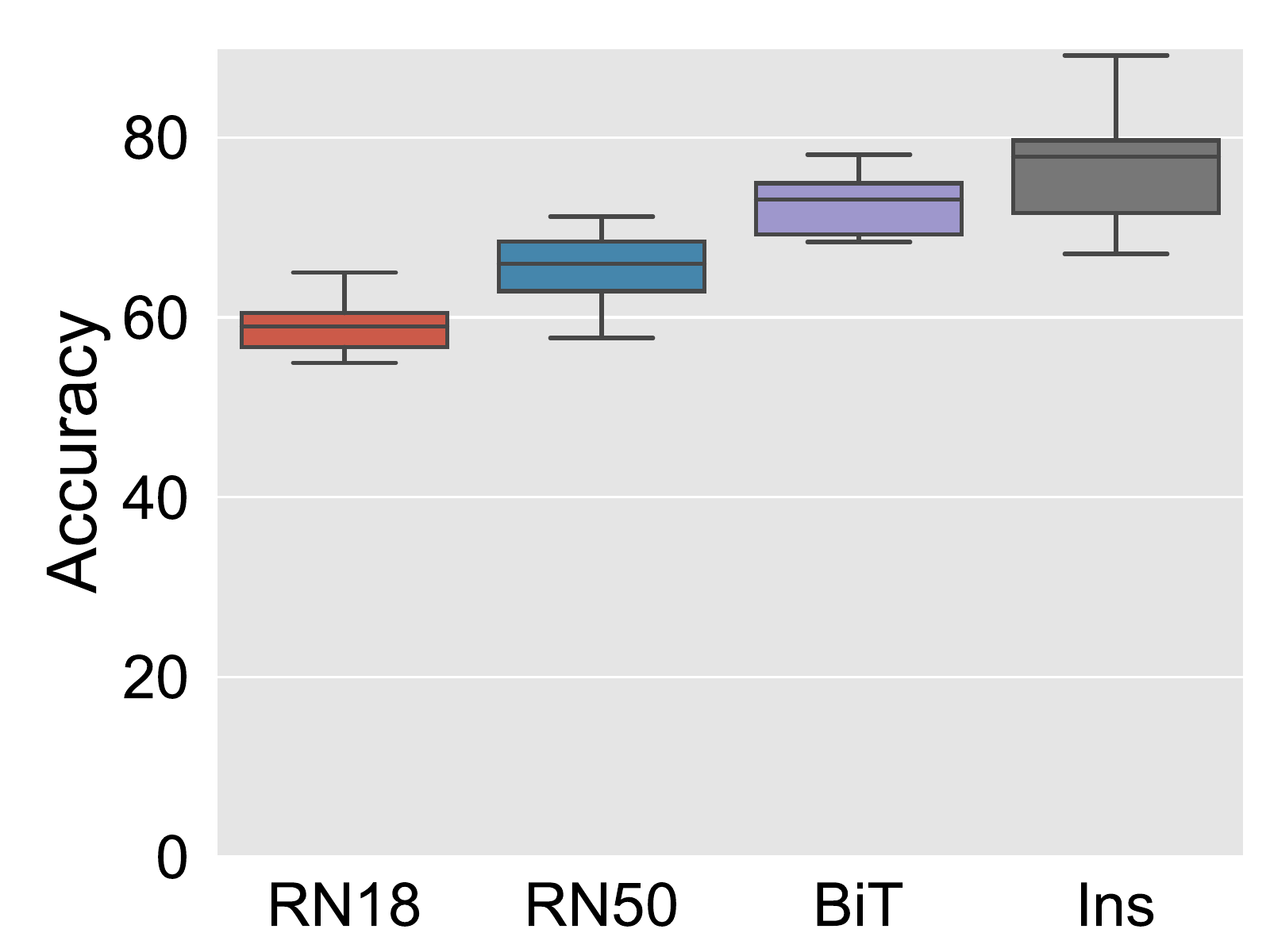}
\caption{VP: STL10}
\label{figure:ablation_k_stl10-prompt}
\end{subfigure}
\caption{The performance gap between different $k$ on different datasets.
The x-axis represents different models.
The y-axis represents the accuracy gap.}
\label{figure:ablation_k}
\end{figure*}

\begin{figure*}[!t]
\centering
\begin{subfigure}{0.5\columnwidth}
\includegraphics[width=\columnwidth]{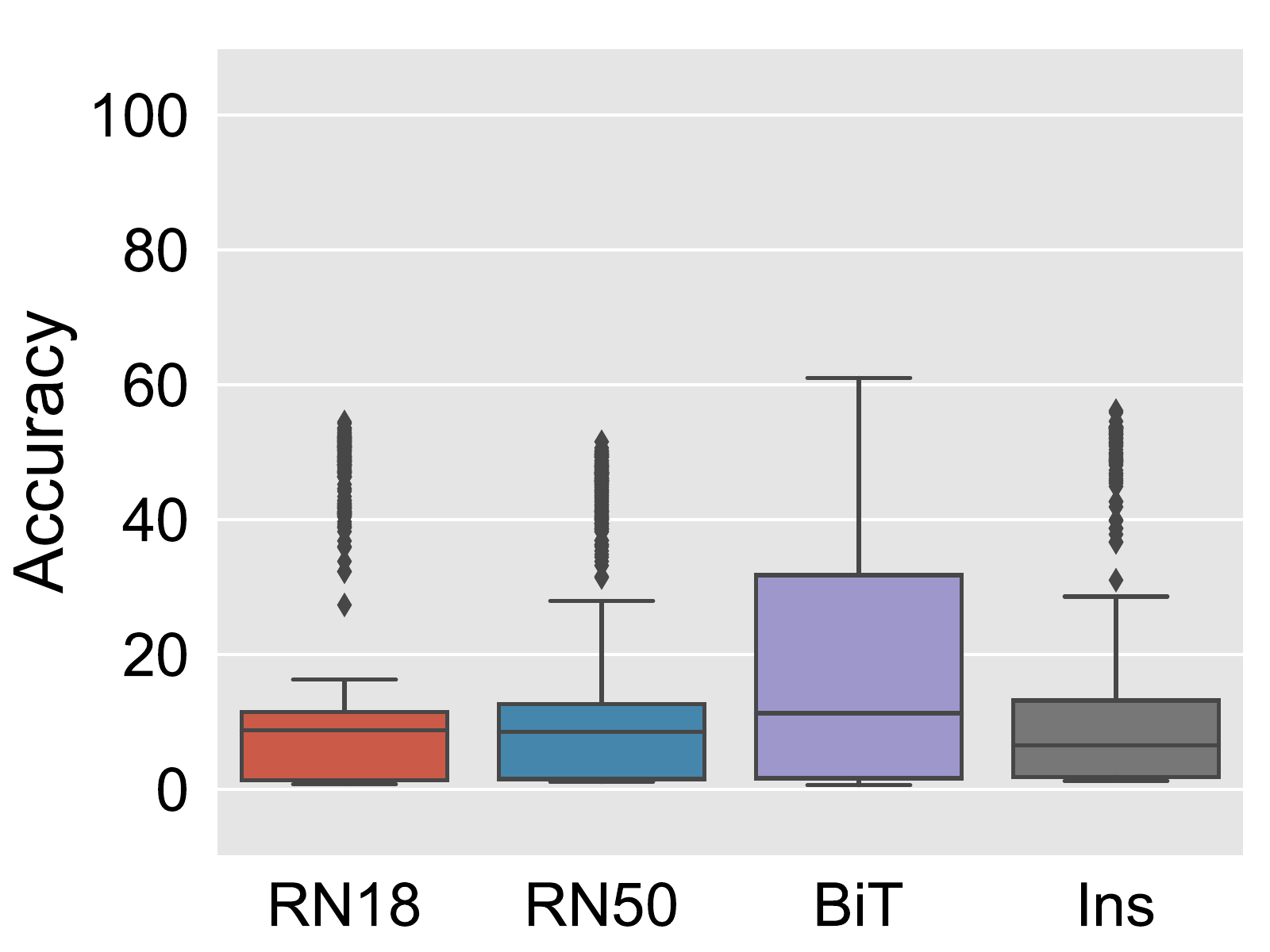}
\caption{RM-VP: CIFAR10}
\label{figure:prompt_cifar10}
\end{subfigure}
\begin{subfigure}{0.5\columnwidth}
\includegraphics[width=\columnwidth]{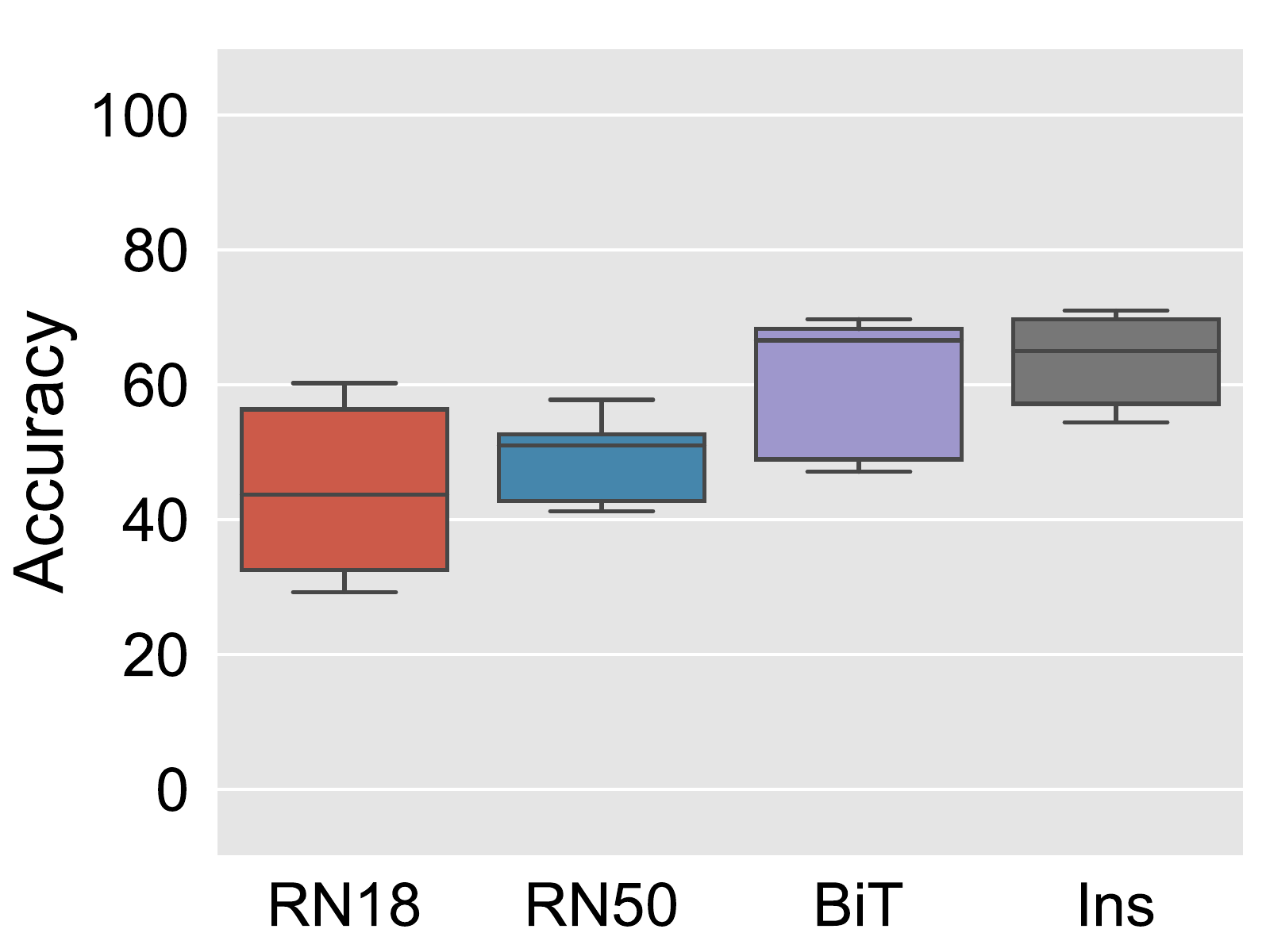}
\caption{\method: CIFAR10}
\label{figure:prompt_cifar10_good}
\end{subfigure}
\begin{subfigure}{0.5\columnwidth}
\includegraphics[width=\columnwidth]{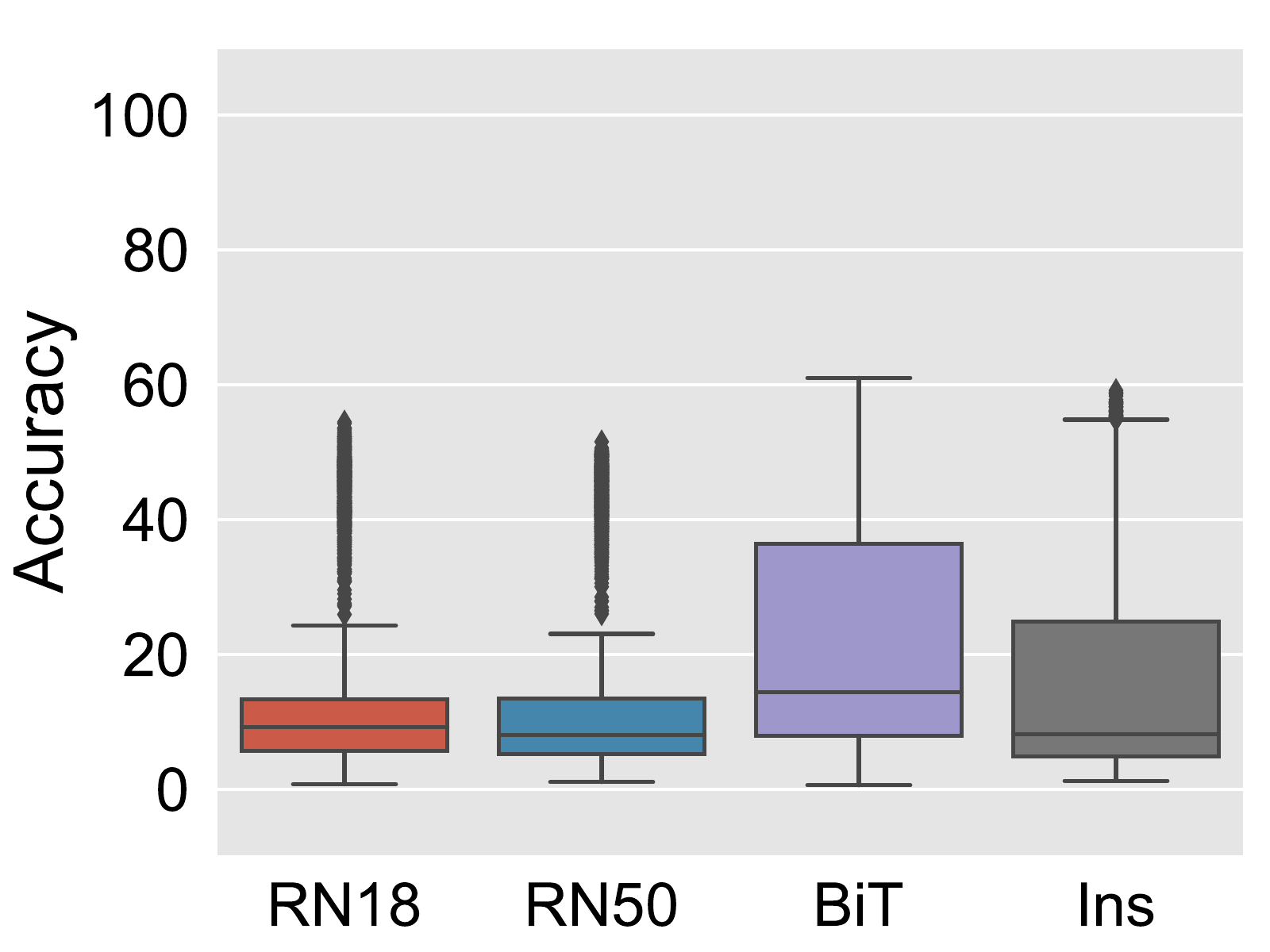}
\caption{RM-VP: CIFAR100}
\label{figure:prompt_stl10}
\end{subfigure}
\begin{subfigure}{0.5\columnwidth}
\includegraphics[width=\columnwidth]{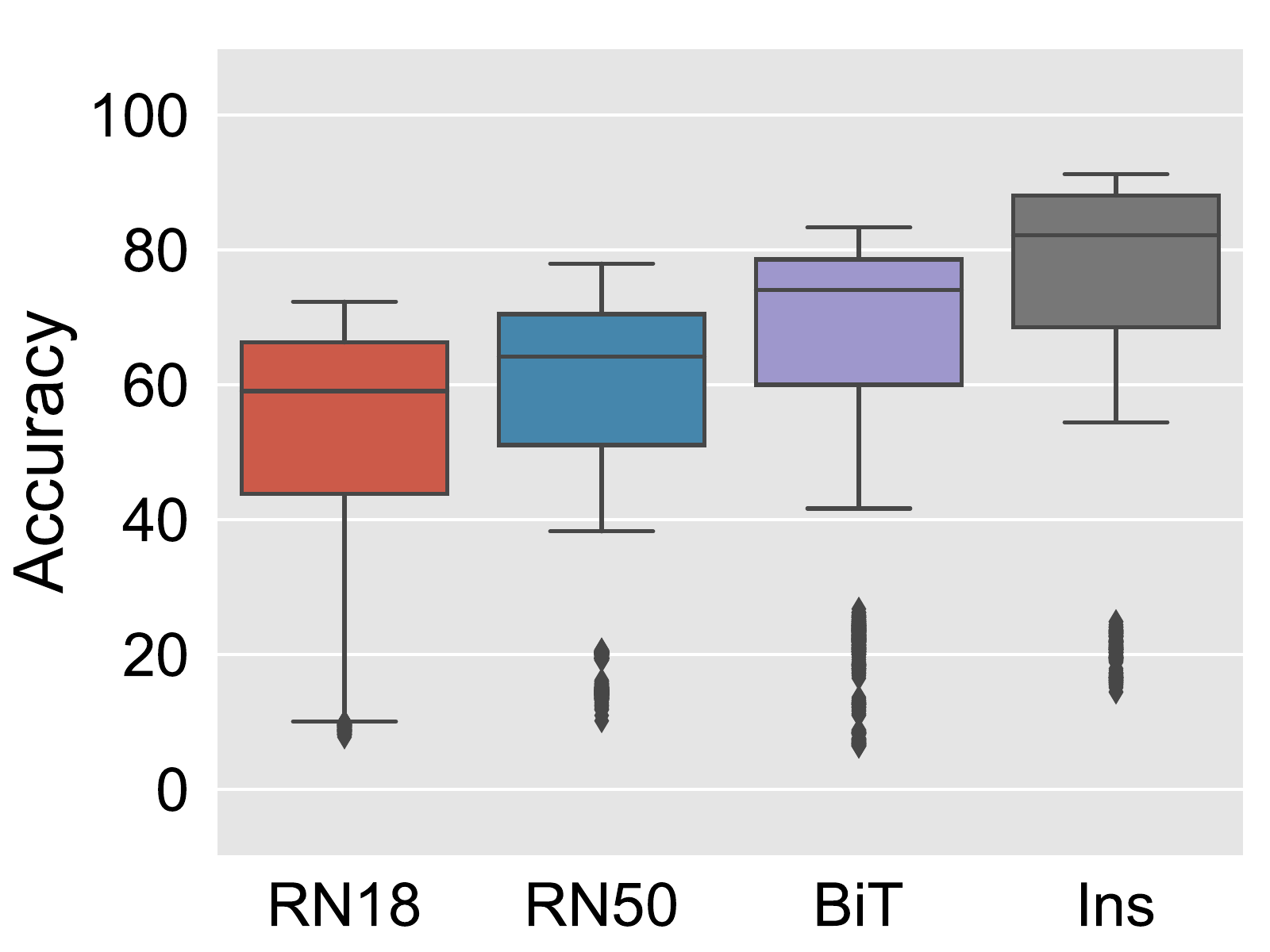}
\caption{\method: CIFAR100}
\label{figure:prompt_stl10_good}
\end{subfigure}
\caption{The performance gap between different prompts on different datasets.
The x-axis represents different models
The y-axis represents the test accuracy.}
\label{figure:prompt_impact}
\end{figure*}

\subsection{Experimental Settings}
\label{subsection:exp-set}

\mypara{Models and Datasets}
Our target model is pre-trained on ImageNet-1k\cite{DDSLLF09}.
We use four different kinds of model structures: ResNet18 (RN18)~\cite{HZRS16}, ResNet50 (RN50)~\cite{HZRS16}, Instagram ResNeXt (Ins)~\cite{XGDTH17}, and BiT-M ResNet50 (BiT)~\cite{KBZPYGH20}.
Note that all of our models are pre-trained on ImageNet-1k.
Specifically, Instagram ResNeXt is first pre-trained on Instagram images with different tags and then fine-tuned on ImageNet-1k.
BiT-M ResNet50 also pre-trained on ImageNet-21k~\cite{DDSLLF09} and then fine-tuned on ImageNet-1k.
For downstream tasks, we evaluate \method on four datasets: CIFAR10~\cite{CIFAR}, CIFAR100~\cite{CIFAR}, F-MNIST~\cite{XRV17}, and STL10~\cite{STL10}.
Our implementation is based on a visual prompt learning framework~\cite{BJSI22}.

\mypara{Baseline Methods}
We compare our methods with the most recent visual prompt learning methods~\cite{CFY21, BJSI22}, where random mapping and frequency-based mapping are adopted.
Therefore, we name the methods as RM-VP and FM-VP.
Further, we compare our methods with traditional fine-tuning, which is denoted as FT.
Note that visual prompt learning and fine-tuning are two different ways to adapt pre-trained models to downstream tasks.

\subsection{Performance of \method}
\label{subsection:performance}

\begin{figure}[!t]
\centering
\begin{subfigure}{0.3\columnwidth}
\includegraphics[width=\columnwidth]{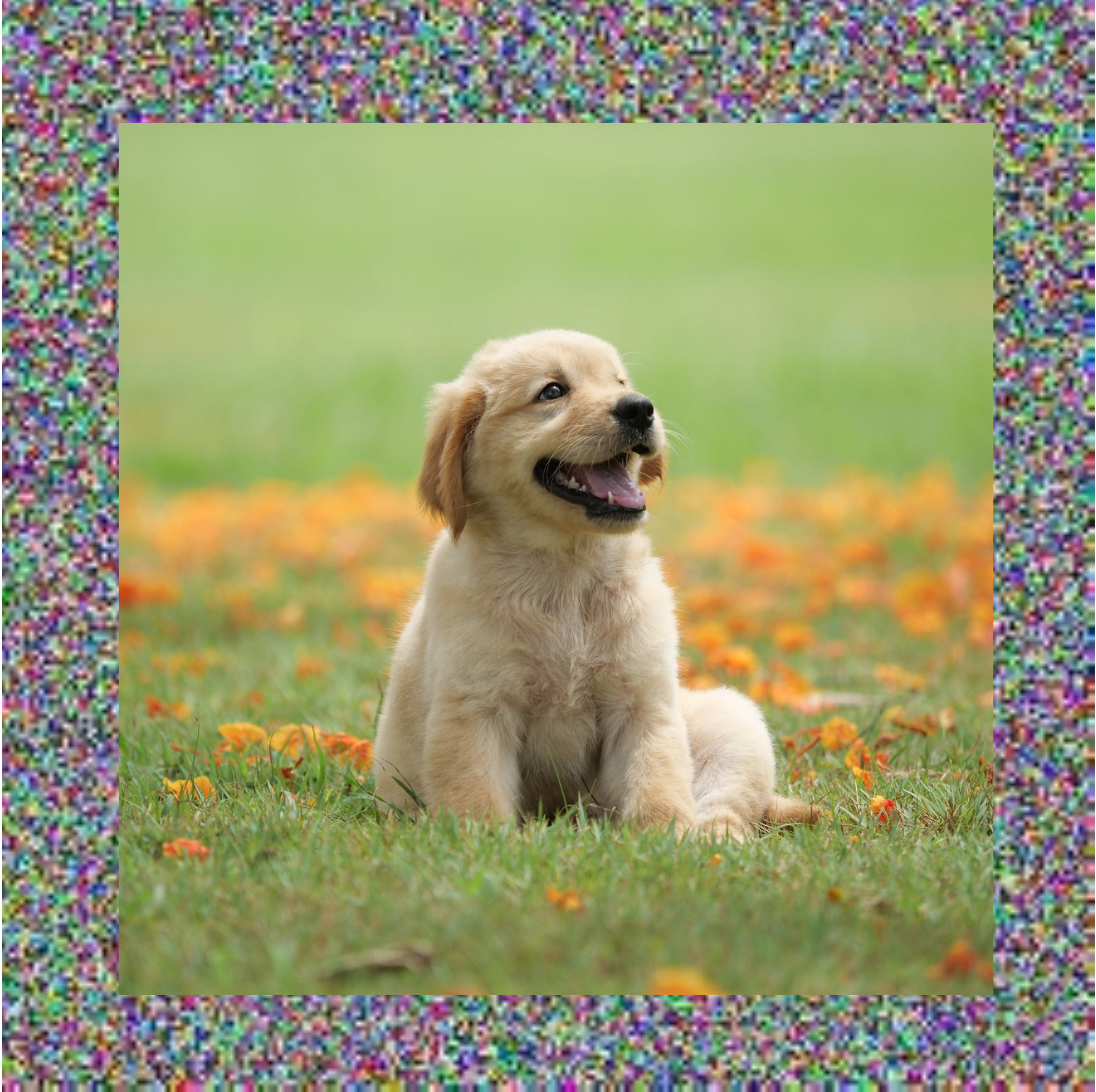}
\caption{Padding}
\label{figure:patching}
\end{subfigure}
\begin{subfigure}{0.3\columnwidth}
\includegraphics[width=\columnwidth]{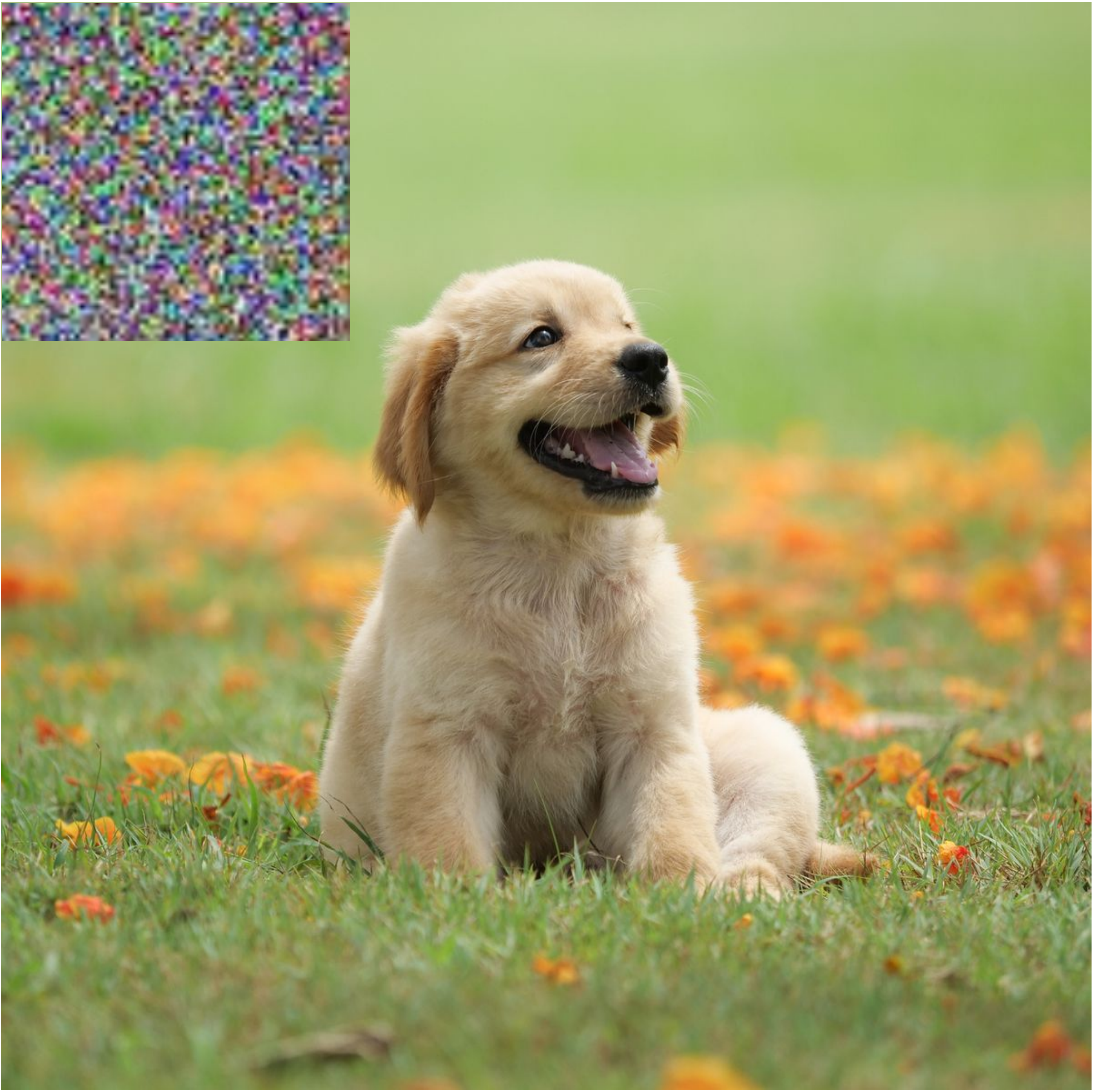}
\caption{Fixed Patch}
\label{figure:fiexed_patch}
\end{subfigure}
\begin{subfigure}{0.3\columnwidth}
\includegraphics[width=\columnwidth]{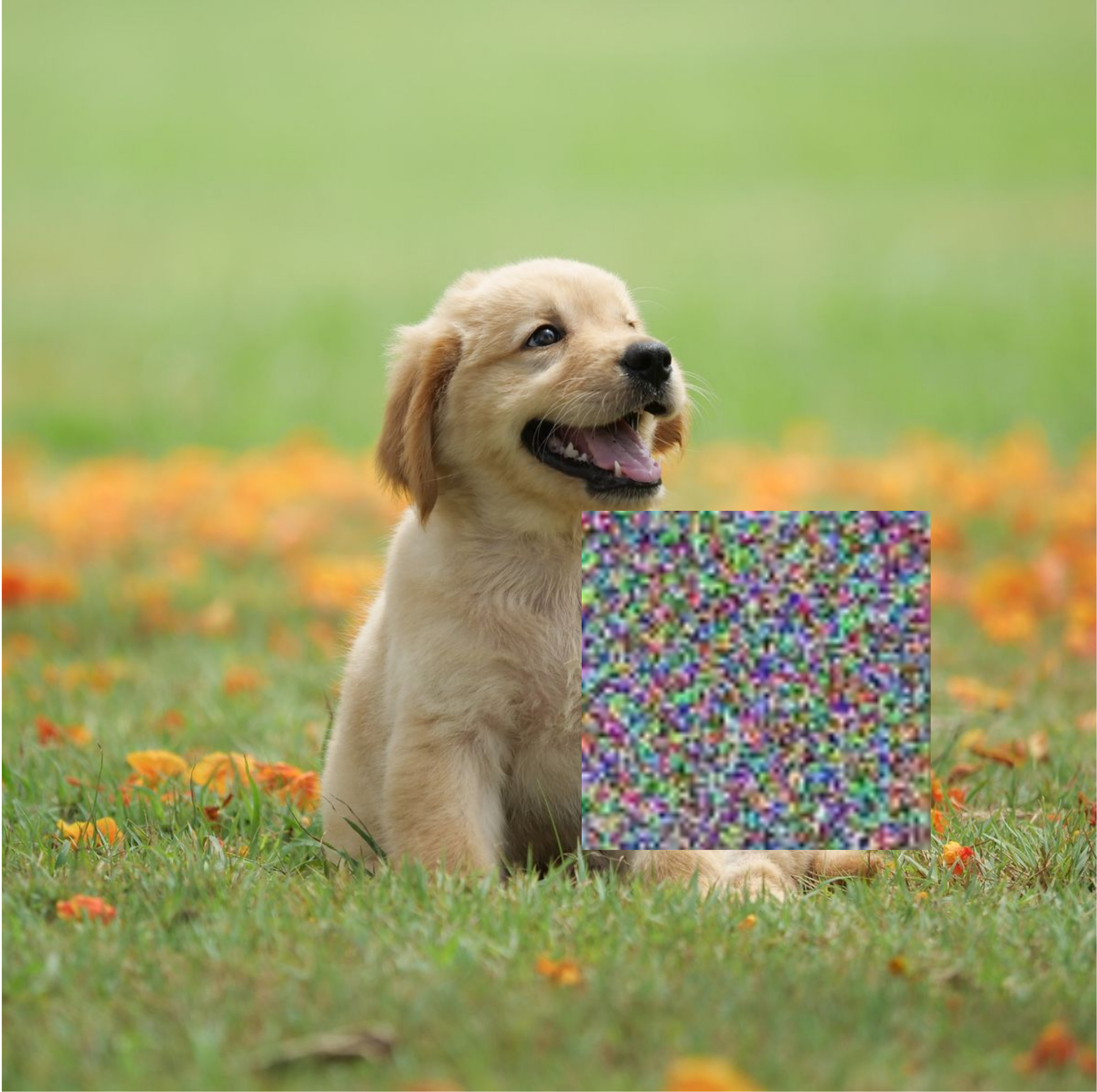}
\caption{Random Patch}
\label{figure:random_patch}
\end{subfigure}
\caption{Different prompts for visual prompt learning.}
\label{figure:patch}
\end{figure}

\begin{figure*}[!t]
\centering
\begin{subfigure}{0.5\columnwidth}
\includegraphics[width=\columnwidth]{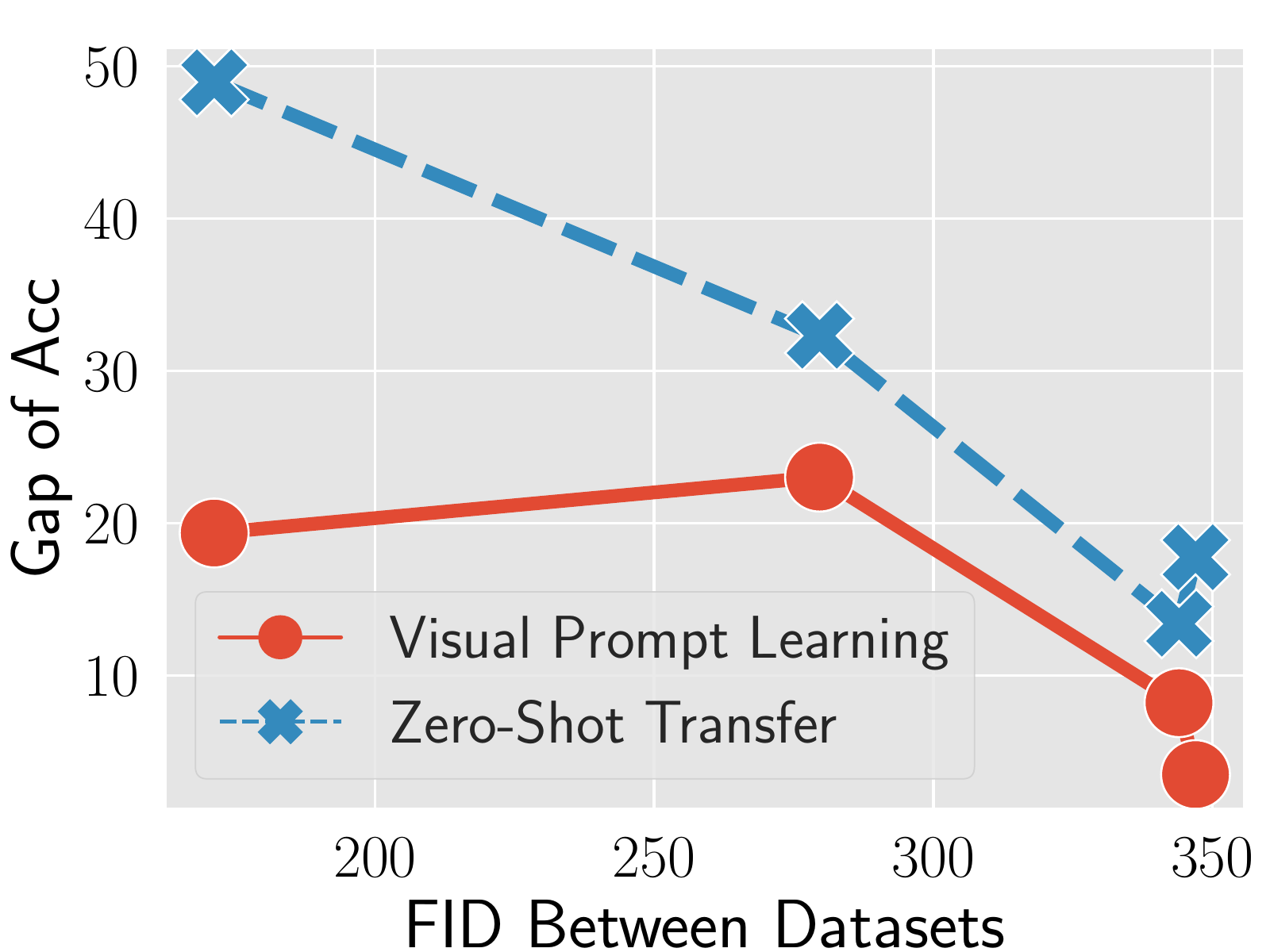}
\caption{ResNet18}
\label{figure:relationrn18}
\end{subfigure}
\begin{subfigure}{0.5\columnwidth}
\includegraphics[width=\columnwidth]{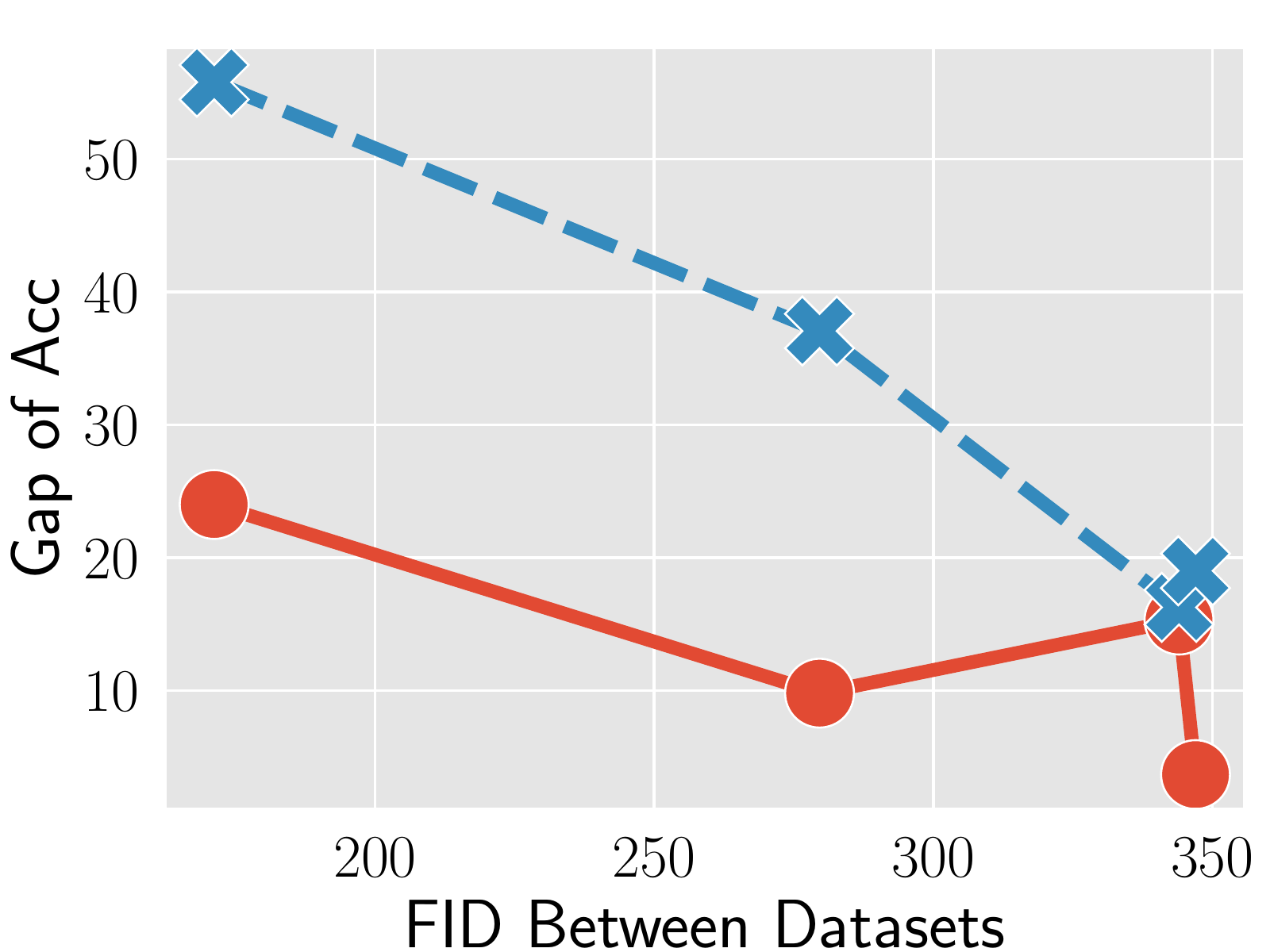}
\caption{ResNet50}
\label{figure:relationrn50}
\end{subfigure}
\begin{subfigure}{0.5\columnwidth}
\includegraphics[width=\columnwidth]{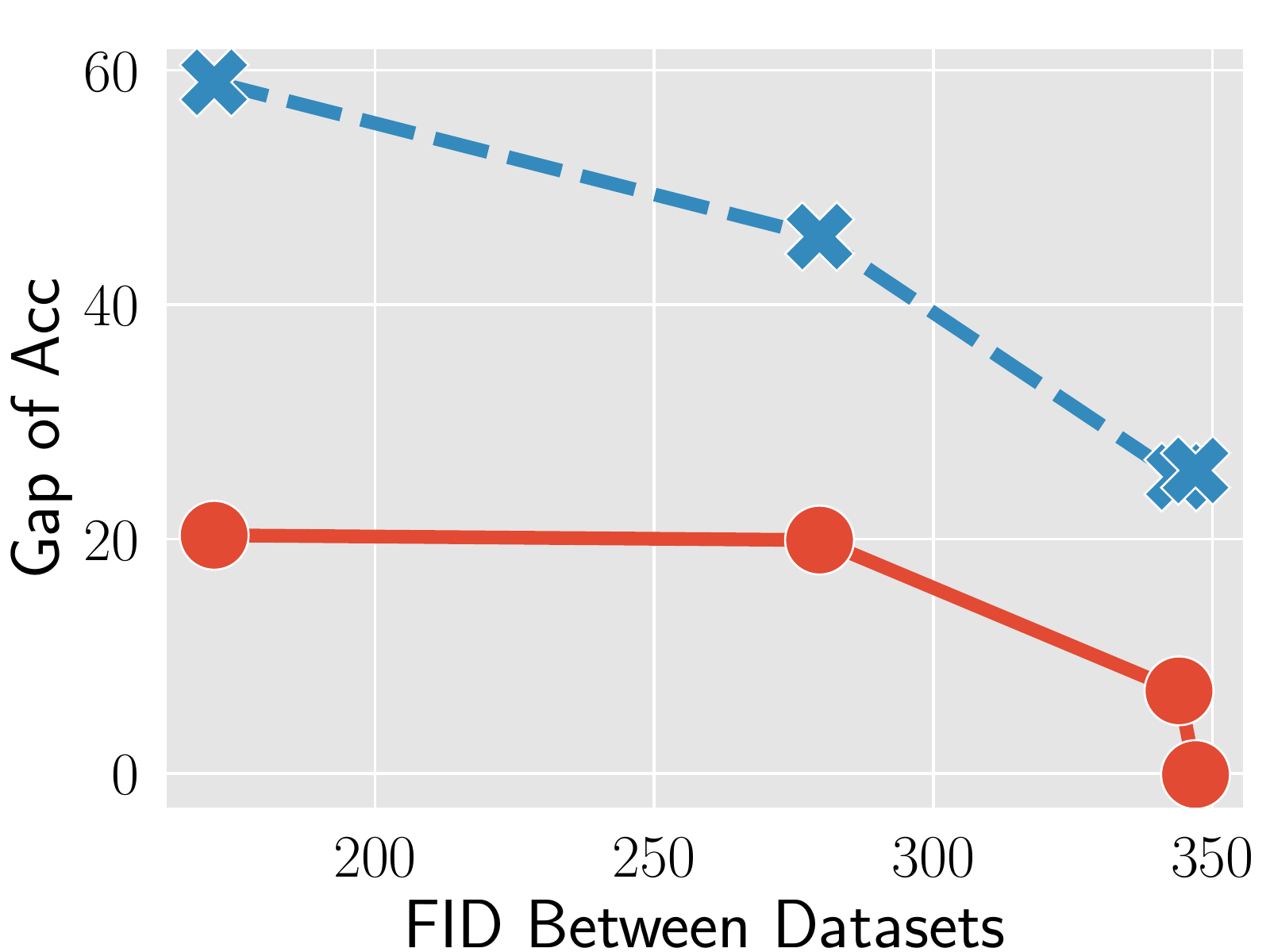}
\caption{BiT-M-ResNet50}
\label{figure:relationrnbit}
\end{subfigure}
\begin{subfigure}{0.5\columnwidth}
\includegraphics[width=\columnwidth]{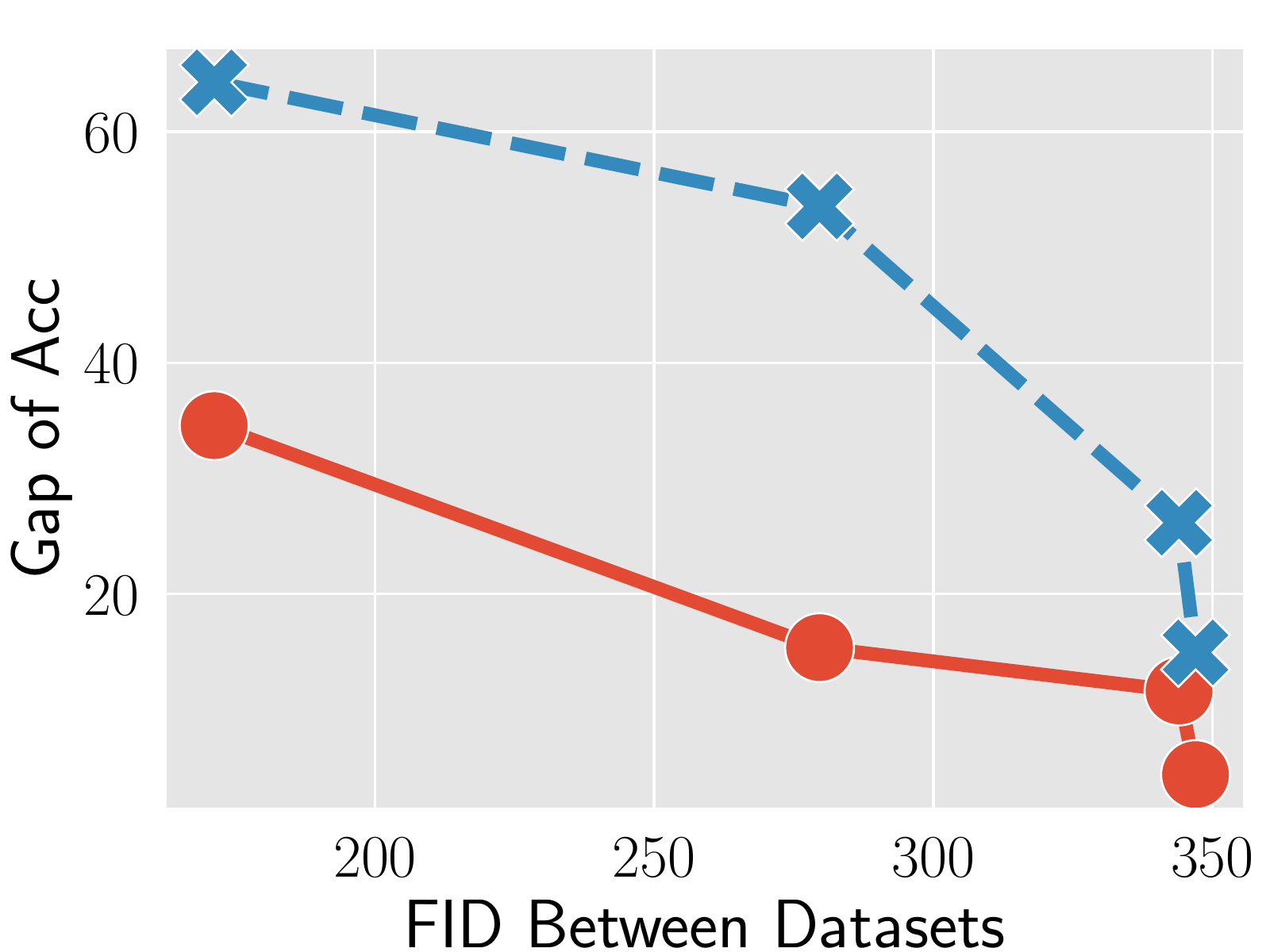}
\caption{Instagram ResNeXt}
\label{figure:relationins}
\end{subfigure}
\caption{The relationship between FID scores and the gap of performance.
The x-axis represents different FID scores between the pre-trained dataset and different downstream datasets (sorted as STL10, CIFAR10, CIFAR100, and F-MNIST).
The y-axis represents the accuracy gap between \method and previous works.
}
\label{figure:relation}
\end{figure*}

In this section, we show that by taking \method as the mapping strategy in visual prompt learning, we can achieve much better performance than RM-VP and FM-VP.
Further, we demonstrate that \method can even achieve good performance in zero-shot transfer without the existence of prompts.

\mypara{Visual Prompt Learning}
We first show the effectiveness of our proposed \method in visual prompt learning.
The results are summarized in \autoref{figure:vp}.
It is shown that \method consistently outperforms existing methods by a large margin.
For instance, when RM-VP/FM-VP can only achieve 0.412/0.174 accuracy on CIFAR10 ResNet18, \method (including \method[-1] and \method[-a]) can achieve the accuracy of 0.570/0.643, which is much higher.
We also find that compared to the intuitive 1-on-1 mapping in \method[-1], taking advantage of mapping multiple indices to one class can have better performance, which preserves more knowledge from the pre-trained model.
For instance, on CIFAR100, \method[-1] can achieve 0.236 accuracy on ResNet50 while the accuracy of \method[-a] is 0.256.
Also, even if the distribution of downstream datasets is far away from pre-trained datasets, for example, on Fashion-MNIST Instagram-ResNeXt, the mapping strategy can still have better performance (0.880 accuracy) than previous works (0.836 accuracy).
This result demonstrates that our \method does not just simply benefit from the clustering results.
Compared to traditional fine-tuning, \method can have better performance on STL10, but is not so good on CIFAR10 and CIFAR100.

Visual prompt learning was proposed to achieve efficient and lightweight knowledge transfer.
It is reasonable to have a worse performance than traditional fine-tuning shown in previous works~\cite{EGS19, CFY21, BJSI22}, as fine-tuning takes more time and computational resources for the transfer.
In the next section, we will introduce zero-shot transfer results, which demonstrate that even without optimization, \method can still better leverage the pre-trained model.

\mypara{Zero-Shot Transfer}
The above section has shown the distinct performance of \method on visual prompt learning.
Now we demonstrate that \method can also be adapted to zero-shot transfer without optimization.
The results are shown in \autoref{figure:zero-shot}.
We observe that in most datasets, such as CIFAR10 and STL10, even without optimization, \method can outperform RM-VP a lot.
This demonstrates that the mapping process is much more important than the prompt optimization process.
We will show more analysis on this conclusion in \autoref{subsection:ablation}.
Also, different from \method of visual prompt learning, the performance of zero-shot transfer depends more on the downstream tasks.
For instance, since STL10 is close to ImageNet-1k, it achieves 0.742 accuracy with \method on Instagram-ResNeXt.
Fashion-MNIST is a totally different dataset from ImageNet-1k, it achieves 0.250 accuracy, which is relatively low but still much higher than random guessing.

\subsection{Ablation Study}
\label{subsection:ablation}

Here, we further conduct ablation studies to explore the effectiveness of our proposed \method.

\mypara{Influence of $k$ in \method[-a]}
We first evaluate the influence of mapping different numbers of classes of the pre-trained model to one downstream class.
As we have demonstrated, \method[-a] can have better performance than \method[-1].
This makes us wonder how the different numbers of target classes, i.e., $k$, impact the overall performance.
\autoref{figure:ablation_k} shows the performance of different $k$ on both visual prompt learning and zero-shot transfer.
With different $k$, the performance varies a lot.
For instance, on CIFAR10 ResNet50, if $k$ is 100, the accuracy reaches 0.516, while it can raise to 0.580 if $k$ is 10.
We also find that our proposed adaptive method \method[-a] can always find a good $k$ for different classes in different datasets.
For instance, on CIFAR10 ResNet18, different $k$ can have an accuracy ranging from 0.496 to 0.651, while \method[-a] achieves 0.642 accuracy.
\autoref{figure:ablation_k} also indicates that $k$ has a larger impact on zero-shot transfer than visual prompt learning.
For instance, on STL10 ResNet50, the performance on zero-shot transfer can range from 0.423 to 0.576, while that of visual prompt learning ranges from 0.577 to 0.712, where the gap is smaller.

\mypara{Mapping VS. Prompt Optimization}
As we have stated, in visual prompt learning, mapping matters more than prompt design.
To better illustrate our conclusion, we conduct experiments to explore the performance of different prompts on both the RM-VP and \method, where different prompts are depicted in \autoref{figure:patch}.
The results are shown in \autoref{figure:prompt_impact}.
Compared to the impact of $k$ shown in \autoref{figure:ablation_k}, the performance gap between good prompts (with a higher performance) and bad prompts (with a lower performance) is larger.
For instance, for CIFAR10 on BiT-M-ResNet50, padding (shown in \autoref{figure:patching}) achieves 0.692 accuracy, while the accuracy of fixed patch (shown in \autoref{figure:fiexed_patch}) is only 0.453.
Also, we observe that even the prompt with the worst performance in \method can have a better performance than the best prompt in RM-VP.
For instance, bad prompt of \method on STL10 on ResNet50 achieves 0.778 accuracy, which is still higher than the 0.473 accuracy on good prompts of RM-VP.
This observation supports our claim that in visual prompt learning, mapping is much more important than the design of prompts.

\mypara{Relationship Between Downstream and Pre-trained Datasets}
Since our methods rely on the semantic relationship between the downstream and pre-trained datasets, it is worthwhile for us to explore the impacts of the relationship between these two datasets.
Therefore, we introduce FID score~\cite{HRUNH17}, which is a metric that measures the similarity between two datasets.
We compute the FID score between ImageNet-1k (as the pre-trained dataset) and the four downstream datasets we have tried.
For visual prompt learning, we calculate the accuracy gap of \method and RM-VP.
For zero-shot transfer, we calculate the accuracy gap of \method and random guess for a fair comparison.
\autoref{figure:relation} shows the results.
We observe that, if the downstream dataset is closer to the pre-trained dataset, e.g., STL10, both zero-shot transfer and visual prompt learning can have better performance.
Also, we find that if the downstream datasets are similar to pre-trained datasets, prompt optimization can largely boost the visual prompt learning performance better.
From \autoref{figure:relation} we also can argue that even though downstream datasets are totally different from the pre-trained dataset, our proposed \method can still have better performance than previous works as the gap is always greater than 0 in all cases.

\section{Conclusion}

In this paper, we focus on exploring the mapping strategies in visual prompt learning.
We first systematically summarize the various paradigms of visual prompt learning.
We demonstrate that mapping is much more important than the design of prompts in visual prompt learning.
We further propose the \method based on semantic closeness.
More specifically, we propose \method[-1] and \method[-a] with 1-on-1 mapping and $k$-on-1 mapping.
To find the best $k$ for \method[-a], we propose adaptive methods to detect the natural gap among similarities.
Extensive experiments show that our \method can have much better performance than previous works.
Moreover, we find that with the mapping strategies in \method, we can take advantage of large visual models for zero-shot transfer tasks without optimizing prompts.
Further experiments show that with \method, the zero-shot transfer can even have better performance than previous training-based methods in visual prompt learning.
It better demonstrates the advantages of our proposed \method.
We also conduct ablation studies to show that mapping matters more than the optimization of prompts by computing the gap of the accuracy of different prompts and different $k$ in $k$-on-1.
We hope that our work will help the community better understand visual prompt learning.
As our proposed \method can achieve that good performance, we hope to provide new insights for the efficient usage of large models with limited computational cost.

\bibliographystyle{plain}
\bibliography{normal_generated_py3}

\end{document}